\documentclass[10pt,twocolumn,letterpaper]{article}

\usepackage{cvpr}              % To produce the CAMERA-READY version
\usepackage{multirow}
\usepackage{comment}
\usepackage{booktabs}
\usepackage{multirow}
\usepackage{array}
\usepackage{graphicx}
\usepackage{makecell}
\usepackage{tabularx}
\usepackage{comment}
\newcommand{\vcell}[2][6.8cm]{%
  \parbox[c][#1][c]{\linewidth}{\centering #2}%
}
\newcommand{\theadcenter}[1]{\multicolumn{1}{c}{\textbf{#1}}}
\usepackage[accsupp]{axessibility}
\definecolor{cvprblue}{rgb}{0.21,0.49,0.74}
\usepackage[pagebackref,breaklinks,colorlinks,allcolors=cvprblue]{hyperref}

\def\paperID{139} % *** Enter the Paper ID here
\def\confName{CVPR}
\def\confYear{2026}

\title{iDiff: Interpretable Difference-aware Framework for Pairwise Image Quality Assessment}

\author{Xinli Yue$^{*}$
\hspace{.1in}
JianHui Sun\thanks{Equal contribution.}
\hspace{.1in}
Tao Shao
\hspace{.1in}
Liangchao Yao
\hspace{.1in}
Fan Xia
\hspace{.1in}
Yuetang Deng\thanks{Corresponding author.}
\\
WeChat, Tencent
}
\begin{document}
\maketitle

\begin{abstract}
Pairwise image quality assessment (IQA) in professional photography requires a model not only to identify the preferred image between two candidates, but also to provide convincing and image-grounded reasoning. In the NTIRE 2026 RAIM challenge, this requirement is further emphasized by jointly evaluating preference prediction and rationale generation. To address this task, we propose \textbf{iDiff}, an \textbf{I}nterpretable \textbf{Diff}erence-aware framework for pairwise image quality assessment. Our method adopts a dual-branch design consisting of an Answer Model and a Thinking Model. The Answer Model performs robust preference prediction by explicitly decomposing each sample into left/right global and local views, followed by content-aware specialization for person and scene images and ensemble-based aggregation across backbones. The Thinking Model focuses on rationale generation and is progressively enhanced with expert-style templates, multi-source quality features, and answer-aware supervision conditioned on the Answer Model prediction. In this way, iDiff jointly models discriminative decision making and structured explanation, improving both robustness and interpretability. Extensive experiments demonstrate the effectiveness of the proposed framework on both accuracy and reasoning-quality metrics. Our method achieved \textbf{first place} in the NTIRE 2026 RAIM challenge, showing the effectiveness of integrating explicit difference modeling with structured multimodal reasoning for pairwise IQA.

%Pairwise image quality assessment (IQA) in professional photography requires not only accurate preference prediction, but also fine-grained and image-grounded reasoning. In the NTIRE 2026 RAIM challenge, models are asked to determine which image in a high-resolution pair is preferred by experts and to generate expert-style rationales for the decision. To address this task, we propose a dual-branch framework that combines a discriminative Answer Model with a reasoning-oriented Thinking Model. The Answer Model reformulates each sample into explicit left/right multi-view inputs, enabling more effective pairwise comparison across global and local views. It is further strengthened by content-aware specialization into person and scene branches and by hierarchical ensemble across backbones. In parallel, the Thinking Model improves rationale generation through progressive instruction design, including template-based regularization, quantitative feature grounding, and answer-aware rationale refinement. In this way, our framework jointly enhances preference prediction and reasoning quality, balancing robustness and interpretability. Extensive experiments demonstrate the effectiveness of the proposed design on both discriminative and reasoning-based evaluation metrics. Our method achieved \textbf{Rank First} in the NTIRE 2026 RAIM challenge, showing the promise of integrating direct decision making with structured multimodal reasoning for pairwise IQA.
\end{abstract}
%,cvciqa,deiqt,step3
\section{Introduction}
Image quality assessment (IQA) plays a fundamental role in computational photography, image restoration, and human-centered visual evaluation~\cite{brisque,nima,musiq,ieit,reiqa}. While conventional IQA research mainly focuses on predicting scalar quality scores~\cite{hyberiqa,maniqa,dacnn,sarque}, recent advances in multimodal large language models (MLLMs)~\cite{llava,instructblip,gpt4v,qwen3vl} have opened up a new direction: enabling models not only to judge image quality, but also to explain their decisions in natural language~\cite{qalign,depictqa,qbench,idetex}. This capability is particularly important for professional photography scenarios, where perceptual preference often depends on subtle trade-offs among sharpness, texture, noise, naturalness, and semantic content.

Compared with absolute score prediction, \emph{pairwise quality comparison} is more aligned with human subjective judgment~\cite{lpips}. In real-world photographic evaluation, annotators often find it easier to decide which image is better than to assign an absolute quality score. Moreover, in the NTIRE 2026 RAIM challenge~\cite{ntire26raim_piqa}, the task goes beyond binary preference prediction and additionally requires the model to generate expert-style reasoning for its decision. This setting introduces two major challenges. First, the model must capture subtle perceptual differences between two visually similar high-resolution images. Second, it must produce coherent and expert-aligned rationales, rather than generic or hallucinated explanations.

Existing IQA methods, including both regression-based approaches~\cite{perceptual,musiq,reiqa,sama} and recent MLLM-based quality assessors~\cite{qboost,groundingiqa,qalign,coinstruct,idetex}, do not fully address this challenge. Traditional discriminative models are typically optimized for direct score prediction and lack interpretability. On the other hand, reasoning-oriented MLLM methods often generate plausible explanations, but their predictions may be less robust under fine-grained pairwise comparison. Motivated by this observation, we argue that accurate preference prediction and high-quality rationale generation should be modeled in a complementary manner rather than by a single branch alone.

\begin{figure*}[ht]
	\centering
	\includegraphics[width=1.0\linewidth]{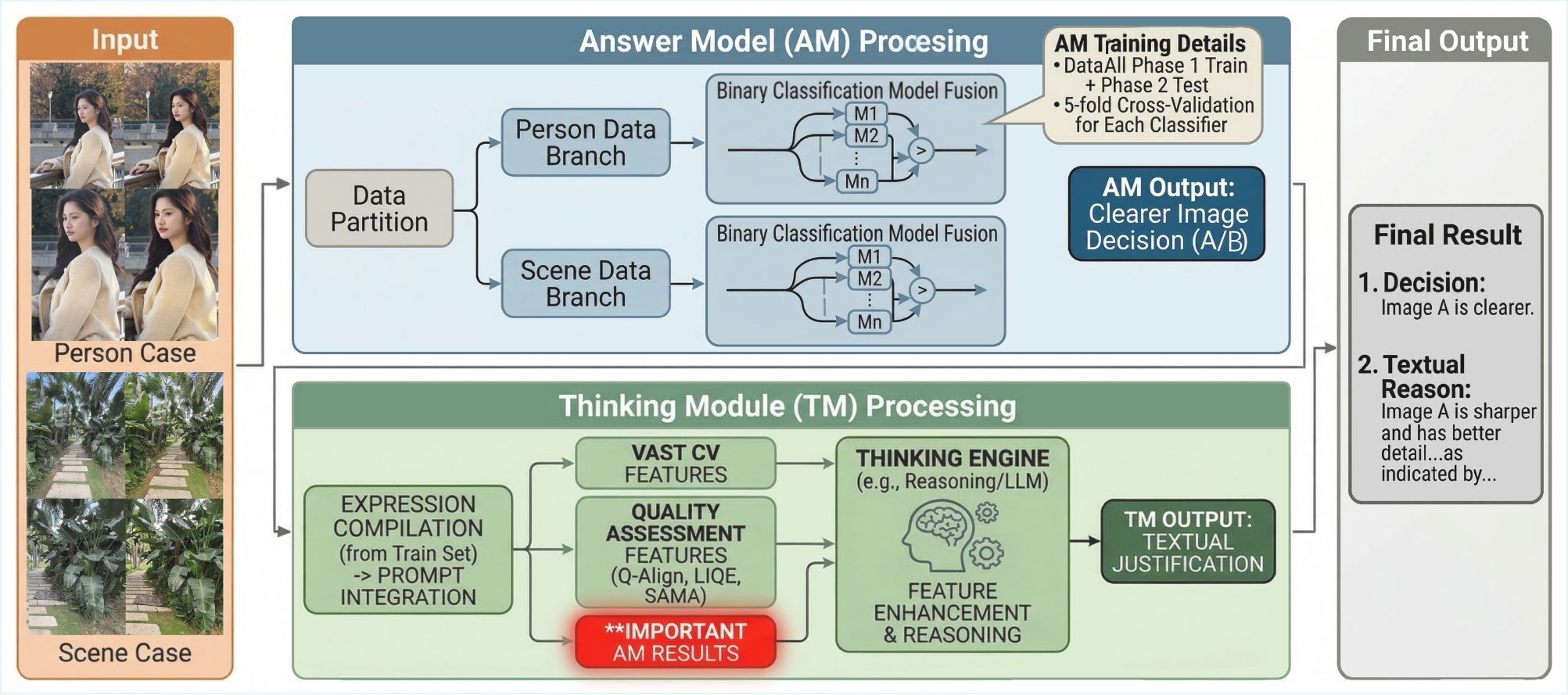}
        %\vspace{-0.2 in}
	\caption{Overview of the proposed dual-branch framework for pairwise image quality assessment. The \textbf{Answer Model (AM)} performs robust preference prediction through content-aware branching and model fusion, while the \textbf{Thinking Model (TM)} generates expert-style textual rationales by integrating prompt templates, quality-related features, and Answer Model outputs. The final output includes both the preferred image decision and its corresponding explanation.}
	\label{fig_overview}
	%\vspace{-0.1 in}
\end{figure*}

To address this challenge, we propose \textbf{iDiff}, an \textbf{I}nterpretable \textbf{Diff}erence-aware framework for pairwise image quality assessment. Our method consists of two collaborative branches: an \textbf{Answer Model} and a \textbf{Thinking Model}. The Answer Model is designed for robust preference prediction. It explicitly decomposes each sample into four aligned views, namely global-left, global-right, crop-left, and crop-right, enabling more effective pairwise difference modeling across both global and local regions. In addition, we adopt content-aware specialization by training separate models for \emph{person} and \emph{scene} images, and further improve robustness through multi-backbone ensemble. In parallel, the Thinking Model is introduced to generate structured and expert-style rationales. It progressively enhances the reasoning ability of an MLLM via template-based regularization, multi-source feature injection, and answer-aware rationale refinement, producing explanations that are more stable, image-grounded, and consistent with the predicted preference.

By combining these two branches, iDiff unifies robust discriminative prediction with interpretable multimodal reasoning. The Answer Model focuses on accurate pairwise decision making, while the Thinking Model explicitly explains the underlying perceptual differences. Extensive experiments on the NTIRE 2026 RAIM benchmark demonstrate that the proposed framework consistently improves both preference prediction and reasoning quality.

Our contributions are summarized as follows:
\begin{itemize}
    \item We propose \textbf{iDiff}, an interpretable difference-aware framework for pairwise image quality assessment, which combines a discriminative Answer Model with a reasoning-oriented Thinking Model.
    \item We design an explicit multi-view pairwise formulation together with content-aware specialization for the Answer Model, enabling robust and accurate preference prediction.
    \item We develop a progressive reasoning enhancement pipeline for the Thinking Model, including template regularization, quantitative feature grounding, and answer-aware rationale refinement, which substantially improves rationale quality and expert alignment.
    \item We achieve strong performance on the NTIRE 2026 RAIM challenge, demonstrating the effectiveness of integrating explicit difference modeling with structured multimodal reasoning for professional pairwise IQA.
\end{itemize}

\section{Related Work}

\paragraph{Image Quality Assessment.}
Early blind image quality assessment (BIQA) methods mainly focus on predicting perceptual quality scores from image features or learned representations. Recent works have started to bridge IQA with vision-language modeling and richer supervision. LIQE~\cite{liqe} formulates BIQA as a vision-language correspondence problem and introduces a multitask learning framework that jointly exploits scene and distortion information, showing that textual task descriptions can help quality prediction. SAMA~\cite{sama} further highlights the importance of jointly modeling global semantics and local details, and proposes a scaling-and-masking data sampling paradigm that preserves multi-scale quality cues within a regular input size. These works demonstrate that modern IQA systems benefit not only from stronger backbones, but also from better problem formulation and data representation.

\paragraph{MLLMs for quality perception and scoring.}
With the rise of MLLMs, recent studies have extended IQA from scalar regression to language-driven quality understanding. Q-Instruct~\cite{qinstruct} improves the low-level visual abilities of multimodal foundation models by constructing instruction-response data grounded in human feedback on clarity, color, and other perceptual attributes. Q-Align~\cite{qalign} further teaches LMMs to perform visual scoring via discrete text-defined quality levels, demonstrating strong performance on IQA, aesthetics, and video quality tasks. DepictQA~\cite{depictqa} moves beyond score prediction and enables detailed language-based image quality evaluation with hierarchical task design and multi-source supervision. These works suggest that language supervision can substantially improve both the perceptual sensitivity and interpretability of quality assessment models.

\paragraph{Comparison-based and explainable IQA.}
Compared with absolute quality prediction, pairwise or open-ended comparison is often more aligned with human subjective judgment. Co-Instruct~\cite{coinstruct} extends LMM-based IQA to open-ended comparative settings, enabling models to answer quality comparison questions and provide free-form reasoning. Compare2Score~\cite{compare2score} further shows that comparative supervision can be used to train an LMM-based visual comparator and then transfer relative judgments to continuous quality estimation. More recently, iDETEX~\cite{idetex} advances detailed and explainable IQA by jointly modeling quality grounding, perception, and description within a unified MLLM framework. Our work is most closely related to this line of research. Different from prior works that mainly target score prediction, open-ended comparison, or generic explainability, we build a dual-branch framework that combines a discriminative Answer Model with a reasoning-oriented Thinking Model for pairwise preference prediction and expert-style rationale generation.

\section{Methodology}
\subsection{Overall Framework}

We propose \textbf{iDiff}, an interpretable difference-aware framework for pairwise image quality assessment. The framework models image preference from two complementary perspectives: direct discriminative prediction and reasoning-guided comparison. As illustrated in Fig.~\ref{fig_overview}, iDiff consists of two collaborative branches, namely an Answer Model and a Thinking Model.

%Our framework addresses pairwise image quality assessment from two complementary perspectives: direct discriminative prediction and reasoning-guided comparison. As illustrated in Fig.~\ref{fig_overview}, the proposed method consists of two collaborative branches, namely an Answer Model and a Thinking Model.

Given an input image pair, where each sample contains both a global comparison view and a local cropped view, we first convert the original concatenated representation into a structured multi-image input. Specifically, each sample is decomposed into four images: global-left, global-right, crop-left, and crop-right. Based on this formulation, the Answer Model directly predicts which candidate image has better perceptual quality, while the Thinking Model generates a structured comparison rationale together with a quality preference decision. The latter follows a rationale-first paradigm, where the model is trained to compare region-level details such as texture, sharpness, noise, and naturalness before arriving at a final judgment.

To better handle the diversity of content, we further divide the data into \emph{person} and \emph{scene} subsets, and train specialized models for each domain. The Answer Model focuses on robust discriminative learning through multi-view decomposition, content-aware specialization, pseudo-label based data expansion, and model ensembling. In parallel, the Thinking Model improves reasoning quality through structured supervision, domain-specific reasoning templates, multi-source feature injection, and answer-aware rationale refinement.

%The two branches play complementary roles in our system. The Answer Model emphasizes prediction accuracy and robustness, benefiting from cross-backbone aggregation, while the Thinking Model improves interpretability by explicitly modeling the comparison process. Together, they form a unified framework that combines the strengths of discriminative classification and reasoning-based assessment for pairwise IQA.

\subsection{Answer Model}
\begin{figure*}[ht]
	\centering
	\includegraphics[width=1.0\linewidth]{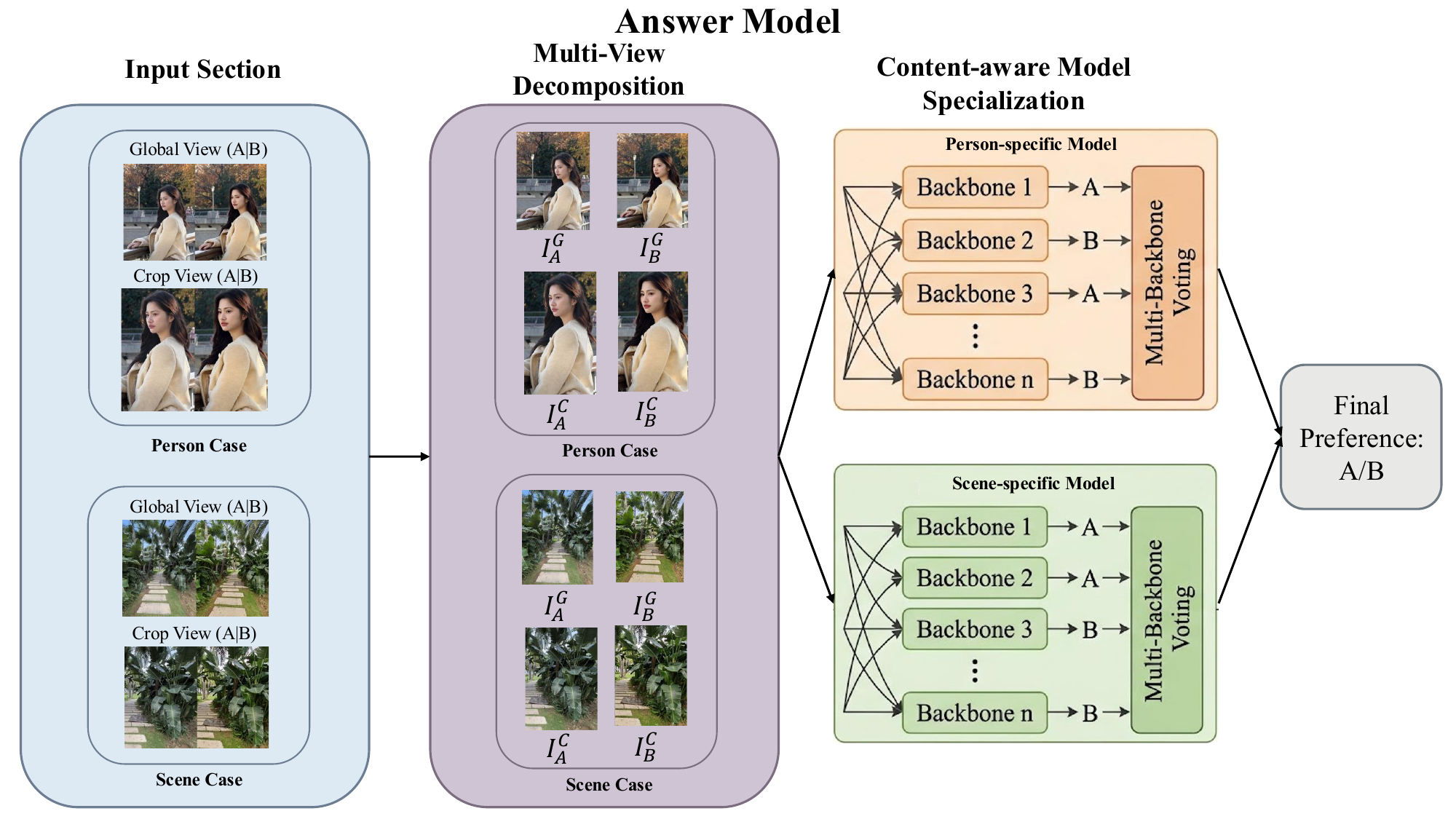}
        %\vspace{-0.2 in}
	\caption{Framework of the proposed Answer Model. The original paired input is reformulated into four aligned views, i.e., global-left, global-right, crop-left, and crop-right, enabling explicit pairwise comparison. The model further adopts content-aware specialization with separate person and scene branches, and uses multi-backbone voting for robust final preference prediction.}
	\label{fig_answer_model}
	%\vspace{-0.1 in}
\end{figure*}
Our Answer Model is designed as a discriminative predictor for pairwise image quality comparison. Given two candidate images, the model directly predicts whether the left image or the right image has better perceptual quality.

\paragraph{Multi-view Input Decomposition.}
The original data is provided as a concatenated image pair, where the left and right candidates are packed into a single image, and each sample contains both a global view and a cropped local view. Instead of directly using the concatenated representation, we explicitly decompose each sample into four aligned inputs:
\begin{equation}
\{I_A^{G}, I_A^{C}, I_B^{G}, I_B^{C}\},
\end{equation}
where $I_A^{G}$ and $I_B^{G}$ denote the global views of the left and right candidates, and $I_A^{C}$ and $I_B^{C}$ denote their corresponding cropped local views. This decomposition allows the model to separately capture global perceptual consistency and local distortion details, which are both critical for pairwise IQA.

\paragraph{Content-aware Model Specialization.}
We observe that different semantic contents exhibit distinct quality cues. For \emph{person} images, perceptual comparison is often dominated by facial texture, skin smoothness, hair boundaries, and restoration artifacts. In contrast, for \emph{scene} images, the judgment relies more on structures such as buildings, vegetation, text clarity, and global naturalness. Therefore, we split the training data into two subsets, namely \emph{person} and \emph{scene}, and train separate Answer Models for each subset. This content-aware specialization reduces intra-class heterogeneity and improves sensitivity to domain-relevant distortions.
\vspace{-8pt}

\paragraph{Multi-backbone ensemble.}
We further ensemble multiple backbone architectures. Different backbones provide complementary inductive biases in modeling texture, edge sharpness, and artifact patterns. Therefore, their fusion consistently improves generalization over any single model. Our final Answer Model prediction is obtained by combining inter-backbone model fusion.

\subsection{Thinking Model}

\begin{figure*}[ht]
	\centering
	\includegraphics[width=1.0\linewidth]{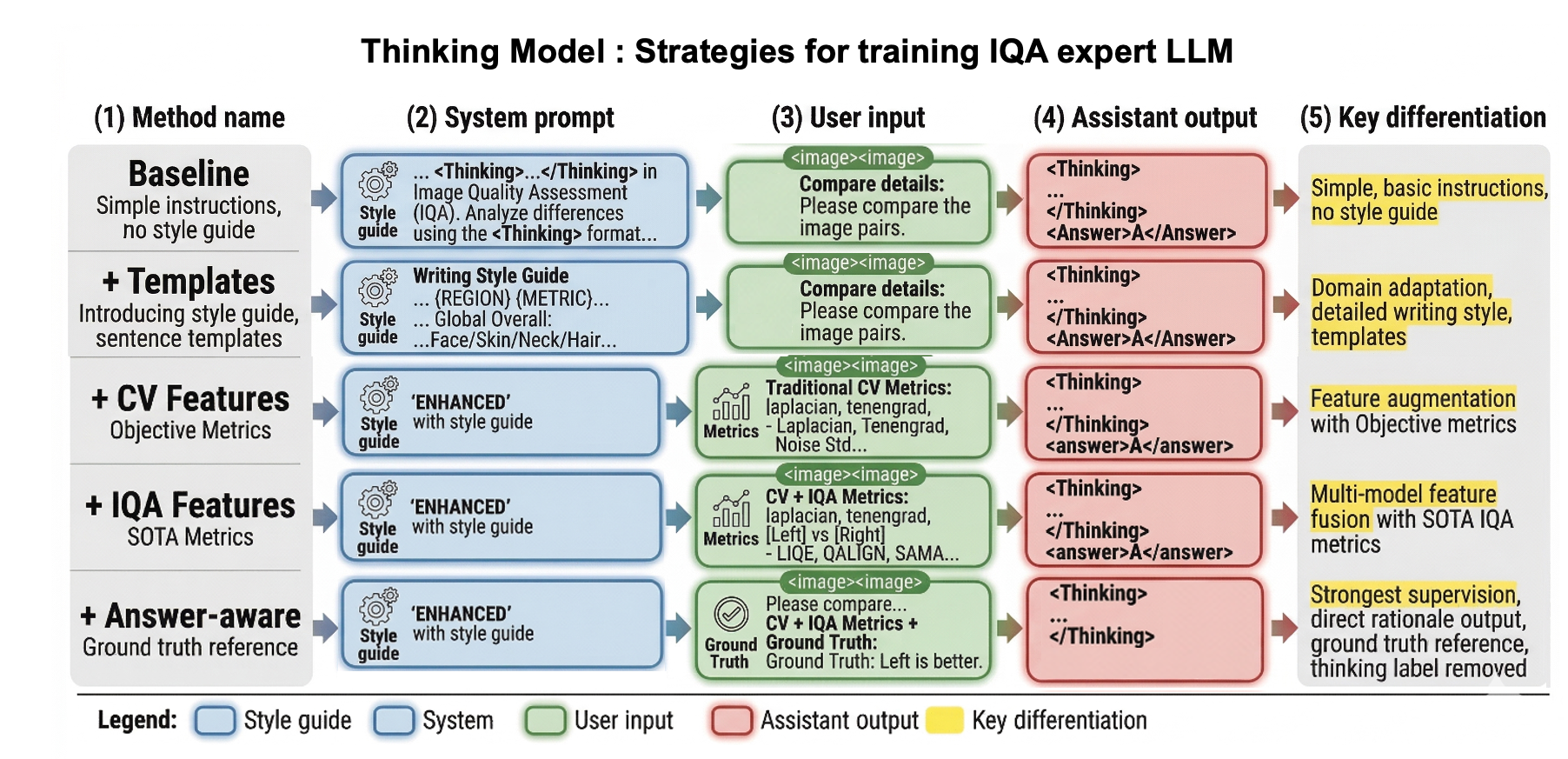}
        %\vspace{-0.2 in}
	\caption{Overview of the progressive instruction design for the proposed Thinking Model. Starting from a baseline rationale-generation setup, we progressively enhance the training strategy with expert-style templates, traditional CV features, learned IQA features, and answer-aware supervision. For each stage, the figure illustrates the corresponding method variant, system prompt, user input, assistant output, and key differentiation. This progressive design improves the reasoning quality of the multimodal large language model and enables more structured, image-grounded, and expert-aligned explanations for pairwise image quality assessment.}
	\label{fig_thinking_model}
	%\vspace{-0.1 in}
\end{figure*}

To enhance the reasoning capability of MLLMs for pairwise image quality assessment, we propose a Thinking Model that explicitly models structured reasoning through prompt engineering and multi-source feature integration.

\paragraph{Structured reasoning supervision.}
Instead of directly predicting the final label, we require the model to generate an intermediate reasoning process in a structured format:
\[
\begin{aligned}
\texttt{<thinking> ... </thinking>} \\
\texttt{<answer>A/B</answer>}
\end{aligned}
\]
The reasoning is decomposed into fine-grained comparisons across multiple perceptual aspects, such as texture, noise, sharpness, and naturalness, together with a final global conclusion. This formulation encourages the model to perform multi-attribute analysis and maintain consistency between local observations and overall judgment. The baseline instruction format follows a rationale-first supervision paradigm, where the model is trained to output both explanations and final decisions.

\paragraph{Template-based reasoning regularization.}
To stabilize reasoning outputs, we introduce expert-style reasoning templates. These templates constrain the model to produce 4--7 concise comparison lines, each focusing on one region and one metric, followed by a mandatory global summary line. Moreover, we design separate templates for different semantic domains. For \emph{person} samples, the templates emphasize facial details such as skin texture, hair edges, and clothing regions, while for \emph{scene} samples, they highlight background texture, architectural structures, foliage, and edge transitions. This domain-aware template design reduces reasoning variance and aligns the outputs with expert-style IQA analysis.

\paragraph{Multi-source Feature Injection.}
To provide the model with additional quantitative cues, we augment the textual input with explicit quality-related features. These include traditional computer vision statistics, such as Laplacian, Tenengrad, noise standard deviation, high-frequency ratio, edge density, entropy, exposure-related ratios, colorfulness, and mean brightness, as well as learned IQA-oriented metrics, including LIQE~\cite{liqe}, Q-Align~\cite{qalign}, and SAMA~\cite{sama}. The features are formatted per image pair and per crop, and then appended to the user prompt as structured auxiliary evidence. This feature-aware prompting enables the model to ground its reasoning in both visual observation and numerical measurements, improving sensitivity to subtle quality differences.
\paragraph{Answer-aware Instruction Tuning.}
We further introduce answer-aware supervision by explicitly including the target answer predicted by the Answer Model in the prompt, e.g., \emph{``Ground truth: Left is better.''} Instead of predicting the answer token again, the model is trained to generate only the reasoning process that supports the provided conclusion. This design prevents trivial answer copying and encourages the model to learn a stronger alignment between evidence and decision. In practice, this variant serves as a rationale refinement stage, where the model focuses on producing more coherent and label-consistent explanations. 

\begin{table*}[t]
\begin{center}
\caption{Results of the NTIRE 2026 The 3rd Restore Any Image Model (RAIM): Professional Image Quality Assessment (Track 1).}
\label{tab_rank}
\begin{tabular}{ccccc}
\toprule[1.0pt]
Rank & Team                       & Phase 2 Score $\uparrow$  & Phase 3 Score $\uparrow$  & Final Score $\uparrow$    \\ \midrule
1 & \textbf{IH-VQA (ours)}     & \textbf{0.6943} & 0.7667          & \textbf{0.7305} \\
2 & VCIP Pi Group              & 0.6826          & \textbf{0.7679} & 0.7253          \\
3 & I\textasciicircum{}2 Group & 0.6490          & 0.7590          & 0.7040          \\
4 & fugui                      & 0.6640          & 0.7237          & 0.6939          \\
5 & LZ                         & 0.6149          & 0.7497          & 0.6823          \\
6 & ongaku                     & 0.6056          & 0.7227          & 0.6641          \\
\bottomrule[1.0pt]
\end{tabular}
%}
\end{center}
\vspace{-0.2 in}
\end{table*}

\section{Experiments}
\subsection{Experimental Setup}
\subsubsection{Dataset}

The NTIRE 2026 RAIM dataset~\cite{raim_piqa} targets \textbf{pairwise image quality assessment} in professional photography scenarios, covering two semantic categories: \textbf{person} and \textbf{scene}. Each sample consists of a pair of high-resolution images, denoted as Image A and Image B, captured by different devices or under different imaging settings. The goal is to predict which image has better perceptual quality.

The dataset contains three subsets. The training set includes 100 image pairs, each annotated with a preference label (\textit{A} or \textit{B}) and a detailed expert-level reasoning description. The Phase-2 validation set contains 102 image pairs without released labels for online evaluation, while the final test set contains 101 image pairs for the final code-submission evaluation.

%The dataset used in the NTIRE 2026 RAIM challenge is designed for pairwise image quality assessment in professional photography scenarios. It mainly covers two semantic categories: \textbf{person} and \textbf{scene}. Each sample consists of a pair of high-resolution images, denoted as Image A and Image B, which are captured by different devices or under different imaging settings. The task is to predict which image in the pair exhibits better perceptual quality.

%The dataset is divided into three parts. The training set contains 100 image pairs. Each training sample includes the image pair, an expert preference label indicating whether Image A or Image B is better, and a detailed expert-level reasoning text explaining the preference. This rationale annotation provides valuable supervision not only for direct classification, but also for reasoning-oriented modeling.

%The Phase-2 validation set contains 102 image pairs without public labels. Participants are required to submit their predictions to the online evaluation server for performance validation. The final test set also contains 101 image pairs and is used for the final code submission evaluation.

%Following the challenge setting, we use the labeled training set as the primary supervised data source. In addition, for the Answer Model, we further exploit the Phase-2 test data through pseudo-labeling and cross-phase aggregation. For the Thinking Model, the expert-provided reasoning annotations in the training set serve as the basis for structured rationale supervision.

\subsubsection{Models}

\paragraph{Answer Model.}
The Answer Model is a discriminative pairwise classifier that predicts the preferred image between two candidates. Each sample is decomposed into four inputs, i.e., global-left, global-right, crop-left, and crop-right, which are encoded by a shared visual backbone for pairwise comparison. To reduce content heterogeneity, we train separate models for face and land samples. We evaluate multiple backbones, including Wide ResNet-50-2~\cite{wideresnet}, EfficientNet-B2~\cite{efficientent}, ConvNeXt-Small~\cite{convnext}, Swin Transformer V2-S~\cite{swinv2}, and MaxViT-T~\cite{maxvit}, and use ensemble for the final prediction.

\paragraph{Thinking Model.}
The Thinking Model is built on MLLMs for rationale generation in pairwise IQA. It takes the image pair and task-specific instructions as input and generates expert-style reasoning, with or without the final answer token depending on the supervision stage. Our main model is Qwen3-VL-8B-Instruct~\cite{qwen3vl}, and we further validate the proposed prompting strategy on MiniCPM-V-4\_5~\cite{minicpm}, GLM-4.6V-Flash~\cite{glm4.6v}, Ovis2.5-9B~\cite{ovis2.5}, and InternVL3\_5-8B~\cite{internvl3.5}.

\subsubsection{Training Parameters}
\paragraph{Answer Model Training.}
We train the Answer Model using AdamW with an initial learning rate of $1\times10^{-4}$ and a weight decay of $1\times10^{-4}$. The batch size is 32, and all models are trained for 10 epochs. Input images are resized to $256\times256$ and randomly cropped to $224\times224$. We use 4 dataloader workers and fix the random seed to 42. 
%In the ablation setting, different backbones are trained in parallel on 8 GPUs.
\vspace{-4pt}

\paragraph{Thinking Model Training.}
We fine-tune Qwen3-VL-8B-Instruct with LoRA-based supervised fine-tuning. The LoRA rank is set to 16, and all linear layers are adapted. Both the vision encoder and aligner are kept trainable. We train for 6 epochs with a per-device batch size of 1, using a learning rate of $4\times10^{-5}$, a weight decay of 0.01, a warmup ratio of 0.03, and a cosine learning rate schedule. Training is conducted in \texttt{bfloat16} precision with DeepSpeed ZeRO-1 on 8 NVIDIA H20 GPUs. 

\subsubsection{Evaluation Metrics}

The challenge evaluates both preference prediction and reasoning quality. Specifically, the final performance is assessed using two types of metrics.
\vspace{-4pt}

\paragraph{Preference Accuracy (ACC).}
ACC measures whether the predicted preference between Image A and Image B matches the expert annotation. It is formulated as a standard binary classification accuracy over the pairwise labels (\textit{A} or \textit{B}).
\vspace{-4pt}

\paragraph{Reasoning Quality.}
The generated reasoning text inside the \texttt{<thinking>} tag is evaluated using a hybrid protocol. First, BLEU-4~\cite{bleu} and {ROUGE-L}~\cite{rouge} are adopted to measure semantic similarity between the generated rationale and the expert-provided reference. %Second, an {LLM-based judge} is used to assess the reasoning quality in terms of logical coherence, hallucination, and image-text alignment.

\begin{table*}[t]
\centering
\setlength{\tabcolsep}{3pt}
\caption{Ablation results of the proposed Answer Model on the Phase-2 validation set. Starting from the baseline, we progressively introduce explicit left/right decomposition, content-aware person/scene specialization, and model ensemble. Unless otherwise specified, all single-model ablation results are based on Swin Transformer V2-S~\cite{swinv2}, while the final ensemble combines multiple specialized models.}
\vspace{-4pt}
\label{tab:ablation_answer}
%\resizebox{\textwidth}{!}{
\begin{tabular}{lccccccc}
\toprule[1.0pt]
Method & Split L/R & Person/Scene & Ensemble & Weighted Voting & Person Acc.  $\uparrow$ & Scene Acc. $\uparrow$ & Overall Acc. $\uparrow$ \\
\midrule
Baseline &  &  &  &  & 0.5658 & 0.6154 & 0.5784 \\
+ Split L/R & \checkmark &  &  &  & 0.7237 & 0.6154 & 0.6961 \\
+ Person/Scene & \checkmark & \checkmark &  &  & 0.7632 & 0.8077 & 0.7745  \\
+ Ensemble & \checkmark & \checkmark & \checkmark & & \textbf{0.8947} & \textbf{0.9615} & \textbf{0.9118}  \\
+ Weighted Voting & \checkmark & \checkmark & \checkmark & \checkmark & \textbf{0.8947} &	0.8846 & 0.8922 \\ 
\bottomrule[1.0pt]
\end{tabular}
%}
\end{table*}

\begin{table*}[t]
\centering
\caption{Comparison between the baseline input formulation and the proposed split L/R design across different backbones on the Phase-2 validation set.}
\vspace{-4pt}
\label{tab:ablation_answer_model}
\begin{tabular}{ccccccc}
\toprule[1.0pt]
\multirow{2}{*}{Model} & \multicolumn{2}{c}{Person Acc. $\uparrow$} & \multicolumn{2}{c}{Scene Acc. $\uparrow$}     & \multicolumn{2}{c}{Overall Acc. $\uparrow$} \\ \cmidrule(l){2-3} \cmidrule(l){4-5} \cmidrule(l){6-7}
                       & Baseline   & Split L/R        & Baseline        & Split L/R       & Baseline    & Split L/R          \\
\midrule
Wide ResNet-50-2~\cite{wideresnet}       & 0.5263     & \textbf{0.6316}  & 0.5769          & \textbf{0.7308} & 0.5392      & \textbf{0.6569}    \\
EfficientNet-B2~\cite{efficientent}        & 0.5658     & \textbf{0.6842}  & \textbf{0.6154} & 0.5000          & 0.5784      & \textbf{0.6373}    \\
ConvNeXt-Small~\cite{convnext}         & 0.5658     & \textbf{0.6184}  & \textbf{0.6923} & 0.5769          & 0.5980      & \textbf{0.6078}    \\
Swin Transformer V2-S~\cite{swinv2}  & 0.5658     & \textbf{0.7237}  & \textbf{0.6154} & \textbf{0.6154} & 0.5784      & \textbf{0.6961}    \\
MaxViT-T~\cite{maxvit}               & 0.5526     & \textbf{0.6184}  & \textbf{0.5000} & \textbf{0.5000} & 0.5392      & \textbf{0.5882}  \\
\bottomrule[1.0pt]
\end{tabular}
\vspace{-0.1 in}
\end{table*}

\subsection{Results on NTIRE 2026 Challenge}

Table~\ref{tab_rank} reports the final results on the NTIRE 2026 RAIM Track 1 challenge leaderboard. Our method ranks 1st place overall with a final score of 0.7305. Specifically, our approach achieves the best Phase 2 Score of 0.6943, which jointly reflects performance in terms of Accuracy, BLEU-4, and ROUGE-L. In Phase 3, which further incorporates LLM-based evaluation alongside Accuracy, BLEU-4, and ROUGE-L, our method achieves a competitive score of 0.7667, leading to the highest overall ranking.

These results demonstrate the effectiveness of our dual-branch design under both evaluation protocols. The Answer Model ensures strong discriminative capability for pairwise preference prediction, contributing to high accuracy, while the Thinking Model enhances the quality of generated rationales, improving text-based metrics such as BLEU-4 and ROUGE-L, as well as the LLM-based score in Phase 3. Their combination enables a balanced optimization of decision accuracy and reasoning quality across different evaluation stages.

%Table~\ref{tab_rank} reports the final results on the NTIRE 2026 RAIM Track 1 challenge leaderboard. Our method achieves the \textbf{1st place} overall with a total score of \textbf{0.6943}. In particular, our approach obtains an accuracy of 0.9118 and achieves the best reasoning score, reaching \textbf{0.1868} in terms of BLEU-4 + ROUGE-L. Although ByteEvalab achieves a slightly higher accuracy of 0.9314, our method delivers substantially stronger reasoning quality, which leads to the best overall ranking under the challenge evaluation protocol.

%These results verify the effectiveness of our dual-branch design. The Answer Model provides strong discriminative performance for pairwise preference prediction, while the Thinking Model significantly improves the quality of the generated rationales. Their combination yields a favorable balance between decision accuracy and explanation quality, which is particularly important in the NTIRE 2026 RAIM setting where both aspects are jointly evaluated. Overall, the challenge results demonstrate that integrating robust preference prediction with structured reasoning is a promising direction for professional pairwise image quality assessment.

\subsection{Ablation Study}

\subsubsection{Answer Model}

Table~\ref{tab:ablation_answer} presents the ablation results of the Answer Model. Starting from the baseline, explicitly splitting the paired image into left/right inputs brings a clear improvement, increasing the overall accuracy from 0.5784 to 0.6961. The gain is more pronounced on the person subset, where accuracy rises from 0.5658 to 0.7237. This confirms that explicit left/right decomposition is more suitable than directly using concatenated inputs, as it better captures relative quality differences.

%Building on the split input design, further introducing content-aware specialization with separate person and scene models leads to another substantial gain. The overall accuracy increases from 0.6961 to 0.7745, while the scene accuracy improves markedly from 0.6154 to 0.8077. This demonstrates that portrait and scenery images exhibit different perceptual cues, and training dedicated models for the two domains effectively reduces content heterogeneity and improves domain-specific discrimination.

Building on this design, introducing content-aware specialization with separate person and scene models further improves performance. The overall accuracy increases from 0.6961 to 0.7745, while the scene accuracy improves significantly from 0.6154 to 0.8077, indicating that domain-specific modeling effectively reduces content heterogeneity and enhances discrimination.

Finally, the average voting ensemble strategy achieves the best results across all subsets, reaching 0.8947 person accuracy, 0.9615 scene accuracy, and 0.9118 overall accuracy. This demonstrates that model fusion provides strong complementarity and further improves robustness.

We further explore a weighted voting strategy, where each model is assigned a weight proportional to its validation accuracy. However, this strategy does not yield additional improvements and slightly reduces the overall accuracy to 0.8922, suggesting that simple equal-weight voting is already sufficiently effective and robust in our setting.

%Table~\ref{tab:ablation_answer} presents the ablation results of the Answer Model. Starting from the baseline, explicitly splitting the original paired image into left/right inputs brings a clear improvement, boosting the overall accuracy from 0.5784 to 0.6961. The gain is especially notable on the person subset, where the accuracy increases from 0.5658 to 0.7237. This result verifies that explicit left/right decomposition is more suitable for pairwise IQA than directly using the concatenated representation, as it enables the model to better capture relative quality differences between the two candidates.

%Finally, the ensemble strategy yields the best performance across all subsets, achieving 0.8947 person accuracy, 0.9615 scene accuracy, and 0.9118 overall accuracy. This confirms that model fusion provides strong complementarity and significantly enhances the robustness of the Answer Model.

To further analyze the impact of the split input design, Table~\ref{tab:ablation_answer_model} compares the baseline and split L/R variants across different backbone architectures. In terms of overall accuracy, all six backbones consistently benefit from explicit left/right decomposition. Although a few backbones show minor fluctuations on individual subsets, the overall trend is highly consistent, indicating that the proposed split L/R formulation is architecture-agnostic and provides a robust improvement across both CNN-based and Transformer-based backbones.

\subsubsection{Thinking Model}

\begin{table*}[t]
\centering
\caption{Progressive ablation of the Thinking Model on the validation set split from the Phase-2 training set. Template regularization, CV features, learned IQA features, and answer-aware prompting each contribute to improved reasoning quality, with the full model achieving the best results.}
\label{tab:ablation_thinking}
\begin{tabular}{lcccccc}
\toprule[1.0pt]
Method & Templates & CV Features & IQA Features & Answer-aware & BLEU-4 $\uparrow$ & ROUGE-L $\uparrow$ \\
\midrule
Baseline &  &  &  &  & 0.0221 & 0.2200 \\
+ Templates & \checkmark &  &  &  & 0.0858 & 0.2825 \\
+ CV Features & \checkmark & \checkmark &  &  & 0.0961 & 0.2994 \\
+ IQA Features & \checkmark & \checkmark & \checkmark &  & 0.1034 & 0.3040 \\
+ Answer-aware & \checkmark & \checkmark & \checkmark & \checkmark & \textbf{0.1161} & \textbf{0.3080} \\
\bottomrule
\end{tabular}
%\vspace{0.1 in}
\end{table*}

\begin{table*}[ht]
\centering
\caption{Comparison between the baseline prompting setup and the proposed Thinking design across different MLLMs on the validation set split from the Phase-2 training set. The proposed design consistently improves reasoning quality for all evaluated models.}
\label{tab:ablation_thinking_model}
\begin{tabular}{ccccc}
\toprule[1.0pt]
\multirow{2}{*}{Model} & \multicolumn{2}{c}{BLEU-4 $\uparrow$} & \multicolumn{2}{c}{ROUGE-L $\uparrow$} \\ \cmidrule(l){2-3} \cmidrule(l){4-5}
                       & Baseline & Thinking        & Baseline  & Thinking        \\ \midrule
MiniCPM-V-4\_5~\cite{minicpm}         & 0.0034   & \textbf{0.0084} & 0.0616    & \textbf{0.1413} \\
GLM-4.6V-Flash~\cite{glm4.6v}         & 0.0169   & \textbf{0.0716} & 0.1786    & \textbf{0.2794} \\
Ovis2.5-9B~\cite{ovis2.5}             & 0.0083   & \textbf{0.0635} & 0.1655    & \textbf{0.2541} \\
InternVL3\_5-8B~\cite{internvl3.5}        & 0.0182   & \textbf{0.0785} & 0.1958    & \textbf{0.2862} \\
Qwen3-VL-8B-Instruct~\cite{qwen3vl}   & 0.0221   & \textbf{0.1161} & 0.2200    & \textbf{0.3080} \\ \bottomrule[1.0pt]
\end{tabular}
\vspace{-0.1 in}
\end{table*}

\begin{table}[ht]
\centering
\caption{Effect of template design on the validation set split from the Phase-2 training set. Template-based prompting significantly improves reasoning quality over the no-template baseline, and person/scene-specific templates perform best.}
\label{tab:template_ablation}
\begin{tabular}{cccc}
\toprule
Method & BLEU-4 $\uparrow$ & ROUGE-L $\uparrow$ \\
\midrule
No template & 0.0221 & 0.2200 \\
Generic template & 0.0801 & 0.2773 \\
Person/Scene templates & \textbf{0.0858} & \textbf{0.2825} \\
\bottomrule
\end{tabular}
%\vspace{-0.2 in}
\end{table}

Table~\ref{tab:ablation_thinking} summarizes the progressive ablation results of the Thinking Model. Starting from the baseline rationale-generation setting, introducing expert-style templates yields the largest improvement, increasing BLEU-4 from 0.0221 to 0.0858 and ROUGE-L from 0.2200 to 0.2825. This highlights the importance of template regularization in stabilizing the output structure and aligning the generated reasoning with expert annotations.

Building on this, adding traditional CV features further improves performance, raising BLEU-4 from 0.0858 to 0.0961 and ROUGE-L from 0.2825 to 0.2994, indicating that low-level image statistics provide useful auxiliary cues for pairwise comparison. Incorporating learned IQA features, including Q-Align~\cite{qalign}, LIQE~\cite{liqe}, and SAMA~\cite{sama}, further boosts performance to 0.1034 BLEU-4 and 0.3040 ROUGE-L, suggesting that these perceptual indicators offer complementary information beyond handcrafted features.

Finally, answer-aware prompting achieves the best results, reaching 0.1161 BLEU-4 and 0.3080 ROUGE-L. Although the gain is smaller than that of template regularization, it consistently refines the reasoning outputs and improves alignment with the target explanations.

%Table~\ref{tab:ablation_thinking} summarizes the progressive ablation results of the Thinking Model. Starting from the baseline rationale-generation setting, introducing expert-style templates yields the largest single-step improvement, increasing BLEU-4 from 0.0221 to 0.0858 and ROUGE-L from 0.2200 to 0.2825. This result shows that template regularization plays a crucial role in stabilizing the output structure and aligning the generated reasoning with the style of expert annotations.

%Based on the template-enhanced setting, adding traditional CV features further improves the reasoning quality, bringing BLEU-4 from 0.0858 to 0.0961 and ROUGE-L from 0.2825 to 0.2994. This indicates that explicit low-level image statistics provide useful auxiliary cues for pairwise quality comparison. When learned IQA features, including Q-Align~\cite{qalign}, LIQE~\cite{liqe}, and SAMA~\cite{sama}, are further incorporated, the performance continues to improve to 0.1034 BLEU-4 and 0.3040 ROUGE-L. This suggests that these learned perceptual indicators offer complementary information beyond handcrafted CV features, helping the model produce explanations that better match expert rationales.

%Finally, answer-aware prompting achieves the best performance, reaching 0.1161 BLEU-4 and 0.3080 ROUGE-L. Although the improvement over the previous stage is more modest than that of template regularization, it consistently refines the reasoning outputs and improves their alignment with the target explanations.

Table~\ref{tab:ablation_thinking_model} further compares the baseline prompting strategy and the full Thinking Model across different vision-language models. The proposed Thinking design consistently improves both BLEU-4 and ROUGE-L for all evaluated backbones. In particular, Qwen3-VL-8B-Instruct achieves the best performance, improving from 0.0221 to 0.1161 in BLEU-4 and from 0.2200 to 0.3080 in ROUGE-L. Similar gains are also observed on MiniCPM-V-4\_5, GLM-4.6V-Flash, Ovis2.5-9B, and InternVL3\_5-8B, demonstrating that the proposed reasoning-oriented design is not tied to a specific MLLMs architecture and generalizes well across different model families.

To further analyze the role of template design, Table~\ref{tab:template_ablation} compares three variants: no template, a generic template, and our domain-specific person/scene templates. Both template-based variants substantially outperform the no-template baseline, confirming that imposing an expert-style structure is highly beneficial for rationale generation. Moreover, the person/scene-specific templates further improve over the generic template, raising BLEU-4 from 0.0801 to 0.0858 and ROUGE-L from 0.2773 to 0.2825. This result suggests that content-aware prompt specialization helps the model focus on more relevant perceptual cues for different semantic domains.

\begin{table}[t]
\centering
\caption{Inference efficiency comparison of the thinking model and the answer model on H20 GPUs. 
For the answer model, the reported results are based on a 5-model ensemble.}
\label{tab:efficiency}
\begin{tabular}{ccc}
\toprule[1.0pt]
Model & Latency (s) & Throughput (image/s) \\
\midrule
Answer Model   & 0.21 & 4.76 \\
Thinking Model & 0.75 & 1.34 \\
\bottomrule[1.0pt]
\end{tabular}
\vspace{-0.2 in}
\end{table}

\subsection{Inference Cost}

Table~\ref{tab:efficiency} compares the inference efficiency of the Answer Model and Thinking Model on H20 GPUs. The Answer Model results are obtained using the full 9-model ensemble. Despite the ensemble design, the Answer Model achieves lower latency (0.21\,s) and higher throughput (4.76 images/s) than the Thinking Model, which requires 0.75\,s and achieves 1.34 images/s. This is mainly because the Answer Model performs direct discriminative prediction, whereas the Thinking Model relies on a multimodal large language model with autoregressive rationale generation. Overall, the Answer Model is more efficient for preference prediction, while the Thinking Model provides complementary interpretability at a higher inference cost.

%\vspace{0.15 in}
\section{Conclusion}

We presented a dual-branch framework for pairwise image quality assessment in professional photography scenarios, which combines a discriminative Answer Model for robust preference prediction and a reasoning-oriented Thinking Model for expert-style rationale generation. By explicitly modeling pairwise inputs with left/right decomposition and content-aware specialization, the Answer Model achieves strong prediction robustness, while the Thinking Model improves explanation quality through structured instruction design, template regularization, feature grounding, and answer-aware rationale refinement. Extensive experiments demonstrate that the proposed framework consistently improves both preference accuracy and reasoning quality. Our method achieved \textbf{first place} in the NTIRE 2026 RAIM challenge, validating the effectiveness of jointly integrating direct decision making and structured reasoning for pairwise IQA.
\clearpage
\bibliographystyle{ieeenat_fullname}
\bibliography{main}
\clearpage
\setcounter{page}{1}
\maketitlesupplementary
\appendix

\section{Analysis of Answer-Aware Conditioning}

The Thinking Model in our framework is designed to generate a rationale that is aligned with the final preference decision. To better understand the effect of answer-aware conditioning, we compare four settings in Table~\ref{tab:answer_aware_analysis}: \emph{w/o Answer-aware}, \emph{Wrong Answer}, \emph{Predicted Answer}, and \emph{Ground-truth Answer}.

As shown in Table~\ref{tab:answer_aware_analysis}, introducing answer-aware conditioning improves rationale quality compared with the version without answer-aware prompting. Specifically, conditioning on the predicted answer improves BLEU-4 from 0.1034 to 0.1161 and ROUGE-L from 0.3040 to 0.3080. When the conditioning signal is further replaced by the ground-truth answer, the performance reaches the best results, with 0.1219 BLEU-4 and 0.3108 ROUGE-L. This indicates that a more reliable answer condition helps the model generate explanations that are more consistent with the reference rationales.

At the same time, conditioning on a wrong answer does not provide comparable gains. Although the ROUGE-L score remains close to the baseline, the BLEU-4 score drops to 0.0958, which is clearly lower than that of conditioning on the predicted answer. This suggests that the improvement does not simply come from providing any answer token in advance. Instead, the quality of the conditioning answer plays an important role in determining the usefulness of the generated rationale.

These results suggest that the main benefit of answer-aware prompting lies in improving decision-rationale alignment and explanation quality, rather than serving as evidence of fully unconstrained reasoning. In other words, the Thinking Model should be more precisely understood as an answer-conditioned rationale generator, whose purpose is to produce coherent and visually grounded explanations for the final decision. This design is especially suitable for pairwise image quality assessment, where the consistency between the preference label and its supporting rationale is important for both interpretability and evaluation.

\begin{table}[t]
\centering
\caption{Effect of answer-aware conditioning on rationale generation.}
\label{tab:answer_aware_analysis}
\begin{tabular}{lcc}
\toprule
Method & BLEU-4 $\uparrow$ & ROUGE-L $\uparrow$ \\
\midrule
w/o Answer-aware & 0.1034 & 0.3040 \\
Wrong Answer       & 0.0958 & 0.3054 \\
Predicted Answer       & 0.1161 & 0.3080 \\
Ground-truth Answer          & \textbf{0.1219} & \textbf{0.3108} \\
\bottomrule
\end{tabular}
\end{table}

\section{Qualitative Analysis}

Tables~\ref{tab:qualitative_cases_face} and \ref{tab:qualitative_cases_land} provide representative qualitative examples on the person and scene categories, respectively. For each case, we present the four input images, the predictions of five Answer Models, the final voting result, and the rationale generated by the Thinking Model. Overall, these examples show that the proposed iDiff pipeline can produce not only accurate final decisions, but also coherent and content-aware explanations that are well aligned with the voting outcomes.

For the \textbf{person} category, the success case shows highly consistent behavior across all five Answer Models, which unanimously prefer image A. The corresponding rationale focuses on face texture, facial sharpness, and hair detail, correctly identifying that image A preserves more natural skin texture and finer high-frequency details, while image B suffers from smearing and detail loss. In the challenging case, although the individual Answer Models are not fully consistent, the final vote still selects image B, and the Thinking Model provides a reasonable explanation centered on sharper facial structures, better texture rendering, and lower facial noise. In the failure case, the pipeline selects image A after a split vote, and the generated rationale still follows plausible visual cues such as facial sharpness, skin texture, and hair detail. This suggests that even when the final prediction is incorrect, the Thinking Model remains strongly grounded in perceptually meaningful evidence rather than producing arbitrary justifications.

For the \textbf{scene} category, the success case again shows unanimous voting, where all five Answer Models prefer image B. The generated explanation consistently emphasizes global sharpness, noise suppression, and texture preservation, especially in foreground vegetation and distant structures. In the challenging case, the Answer Models exhibit disagreement, but the final vote still favors image B, and the Thinking Model attributes this preference to clearer scene structures, better noise control, and richer global textures such as water ripples and building facades. In the failure case, the model still produces a visually grounded explanation, highlighting a trade-off between higher sharpness and stronger artifacts, such as noise in dark regions and black-border artifacts near high-contrast edges. These examples indicate that the proposed framework can capture both global scene fidelity and local artifact patterns, while also revealing its limitations in ambiguous cases where detail enhancement and artifact amplification coexist.

Overall, these qualitative results support two key observations. First, the Answer Model ensemble provides stable decisions in both easy and difficult cases, with unanimous voting often corresponding to clear perceptual differences and split voting reflecting higher ambiguity. Second, the Thinking Model generally produces explanations that are consistent with both the final vote and the visible image content, covering category-relevant cues such as facial texture and hair detail for person images, and sharpness, noise, texture, and artifacts for scene images. This further demonstrates the interpretability of the proposed answer-then-thinking pipeline.

\begin{table*}[t]
\centering
\scriptsize
\setlength{\tabcolsep}{2pt}
\renewcommand{\arraystretch}{1.0}
\caption{Qualitative examples on the person category of the proposed iDiff pipeline.
For each case, we show the four input images arranged in a 2$\times$2 layout, the predictions of five Answer Models, the final voting result, and the reasoning generated by the Thinking Model.}
\label{tab:qualitative_cases_face}
\resizebox{\textwidth}{!}{
\begin{tabularx}{\textwidth}
{
>{\centering\arraybackslash}m{0.05\textwidth}
>{\centering\arraybackslash}m{0.45\textwidth}
>{\centering\arraybackslash}m{0.10\textwidth}
>{\centering\arraybackslash}m{0.05\textwidth}
>{\raggedright\arraybackslash}X
}
\toprule[1.0pt]
\textbf{Case} & \textbf{Inputs} & \textbf{Answer Models} & \textbf{Vote} & \theadcenter{Thinking Model}\\
\midrule

\vcell{Success}
&
\begin{tabular}{@{} c >{\centering\arraybackslash}m{0.13\textwidth} >{\centering\arraybackslash}m{0.13\textwidth} @{}}
& \textbf{\tiny A} & \textbf{\tiny B} \\
\textbf{\tiny G} &
\includegraphics[width=\linewidth]{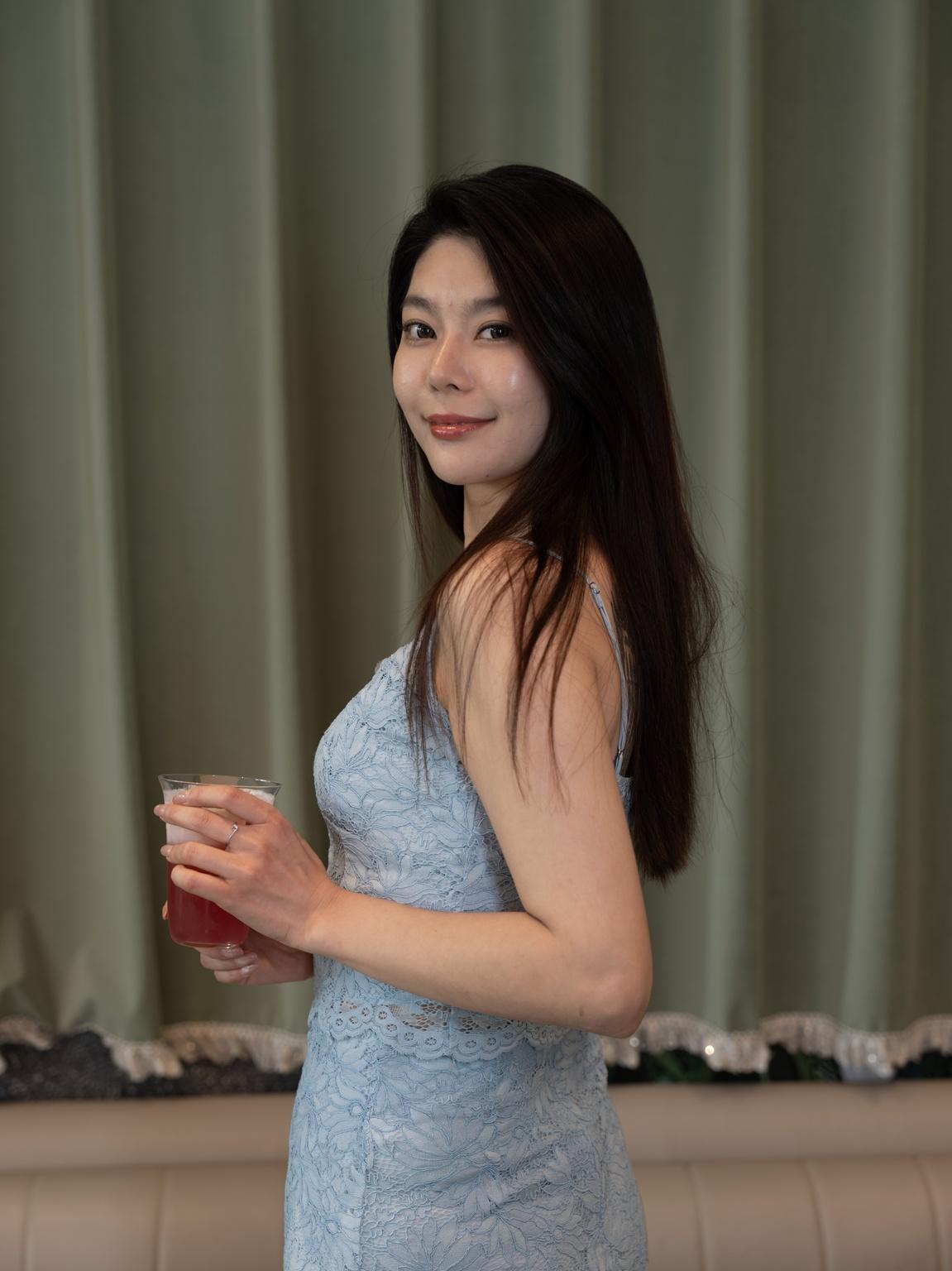} &
\includegraphics[width=\linewidth]{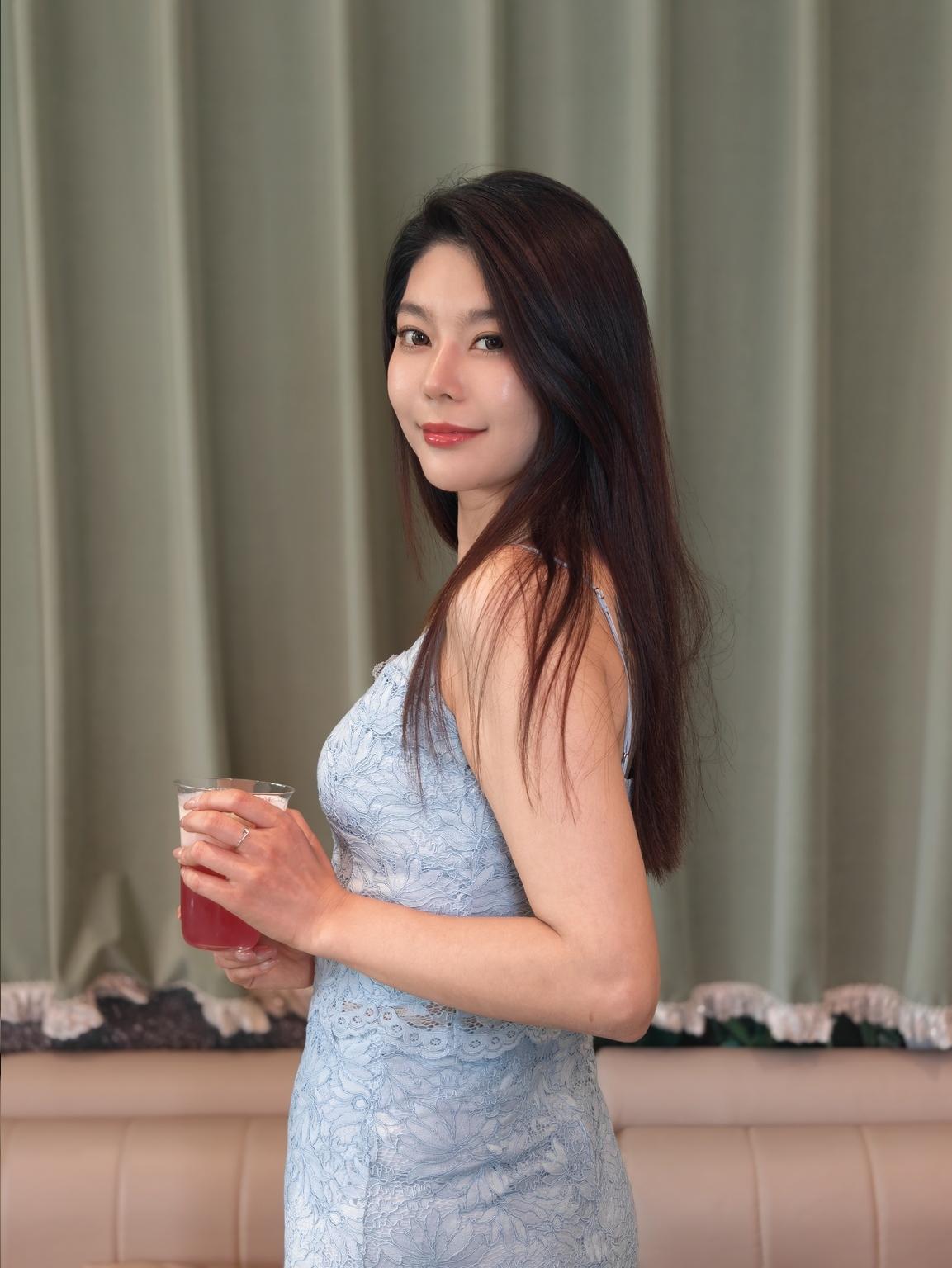} \\
\textbf{\tiny C} &
\includegraphics[width=\linewidth]{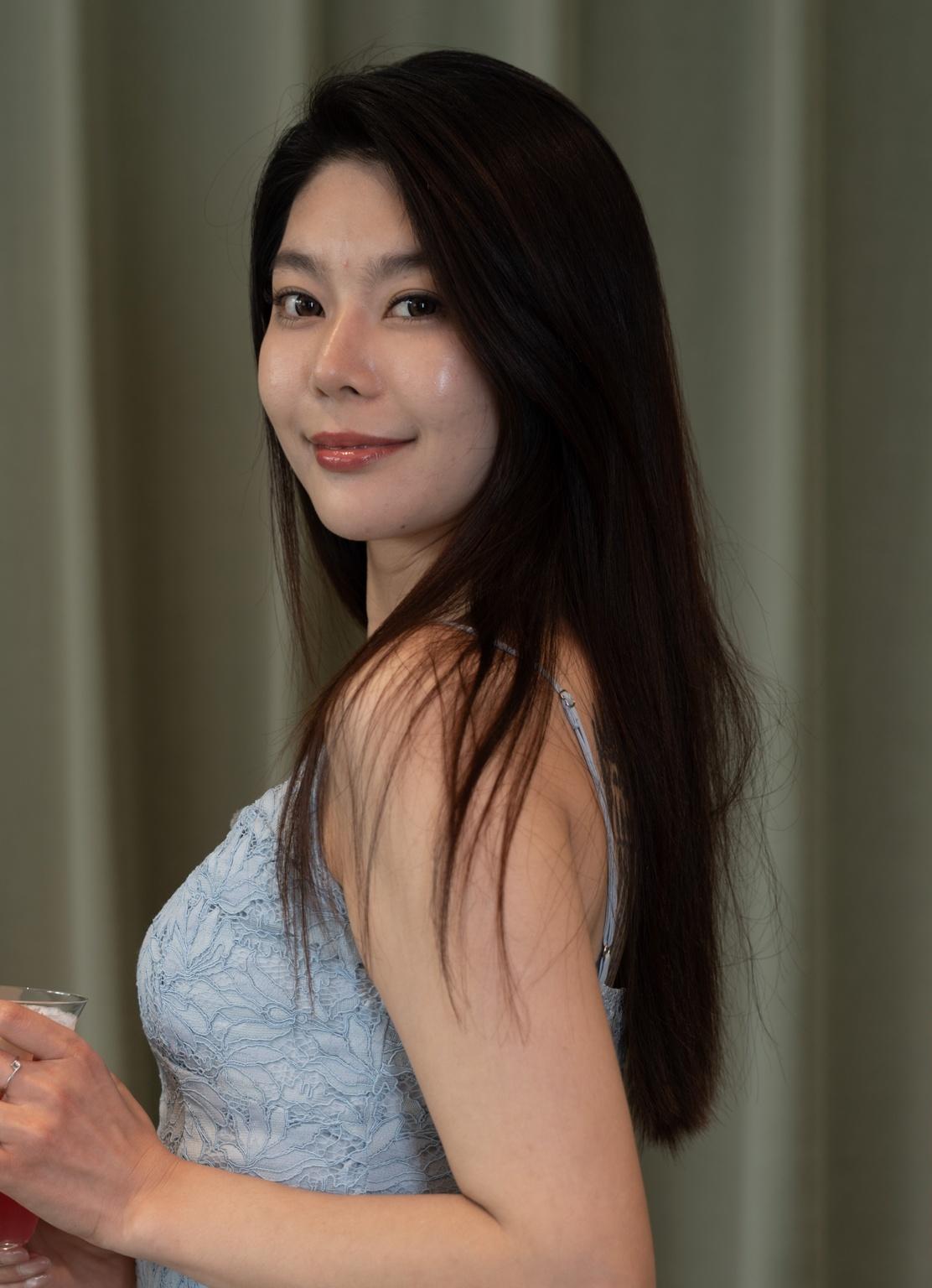} &
\includegraphics[width=\linewidth]{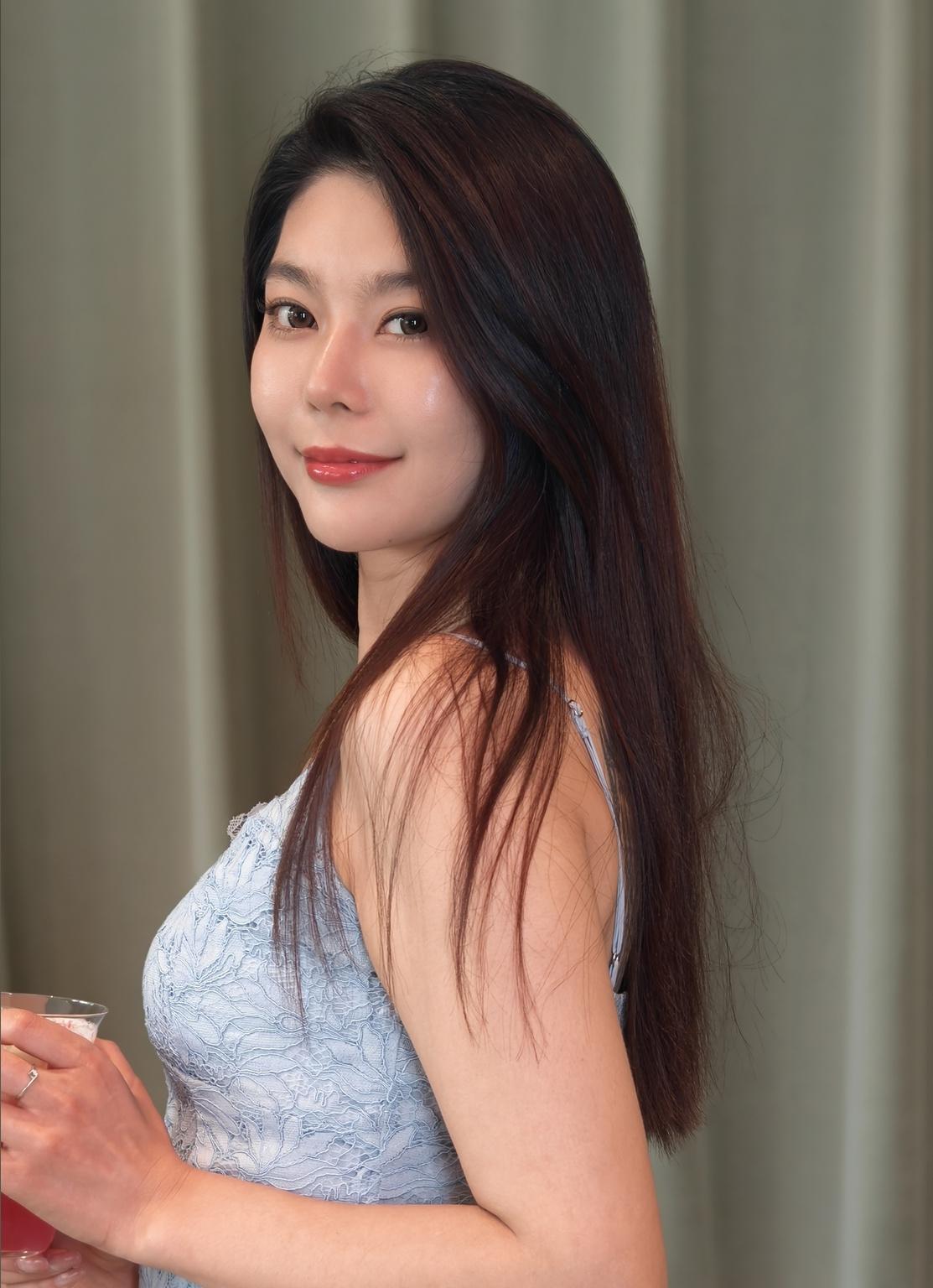}
\end{tabular}
&
\vcell{A / A / A / A / A}
&
\vcell{\textbf{A}}
&
Face Texture: Image A (Left) exhibits more natural and detailed skin texture, while Image B (Right) shows noticeable smearing and loss of fine details, particularly in the mid-to-dark tone regions. Sharpness: Image A (Left) demonstrates superior sharpness in facial features, whereas Image B (Right) appears softer and less defined. Texture: Image A (Left) preserves finer hair strands and texture, while Image B (Right) lacks this level of detail. Sharpness: Image A (Left) maintains overall higher sharpness across the entire image, contributing to better clarity and detail retention.
\\
\midrule

\vcell{Challenging}
&
\begin{tabular}{@{} c >{\centering\arraybackslash}m{0.13\textwidth} >{\centering\arraybackslash}m{0.13\textwidth} @{}}
& \textbf{\tiny A} & \textbf{\tiny B} \\
\textbf{\tiny G} &
\includegraphics[width=\linewidth]{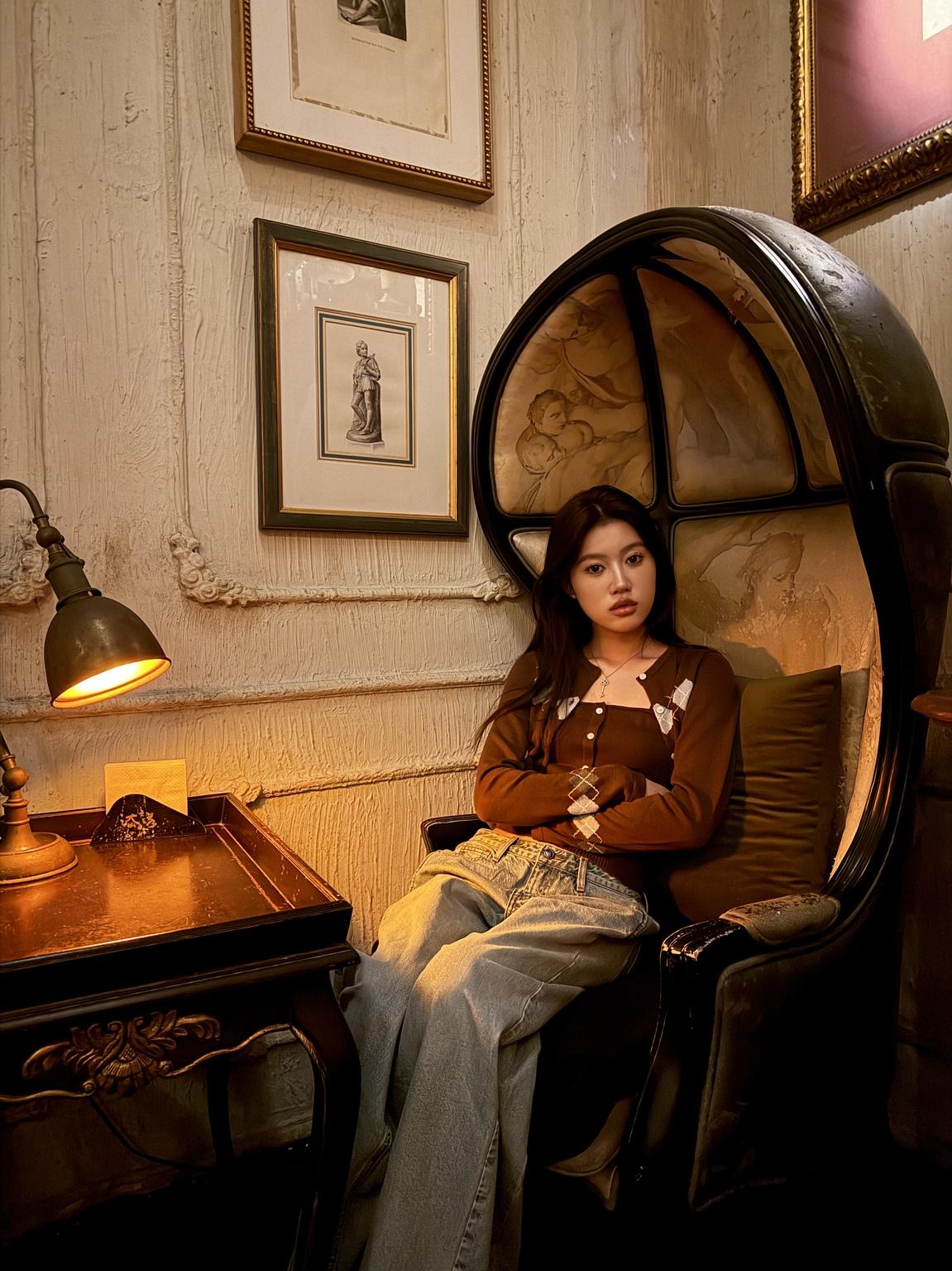} &
\includegraphics[width=\linewidth]{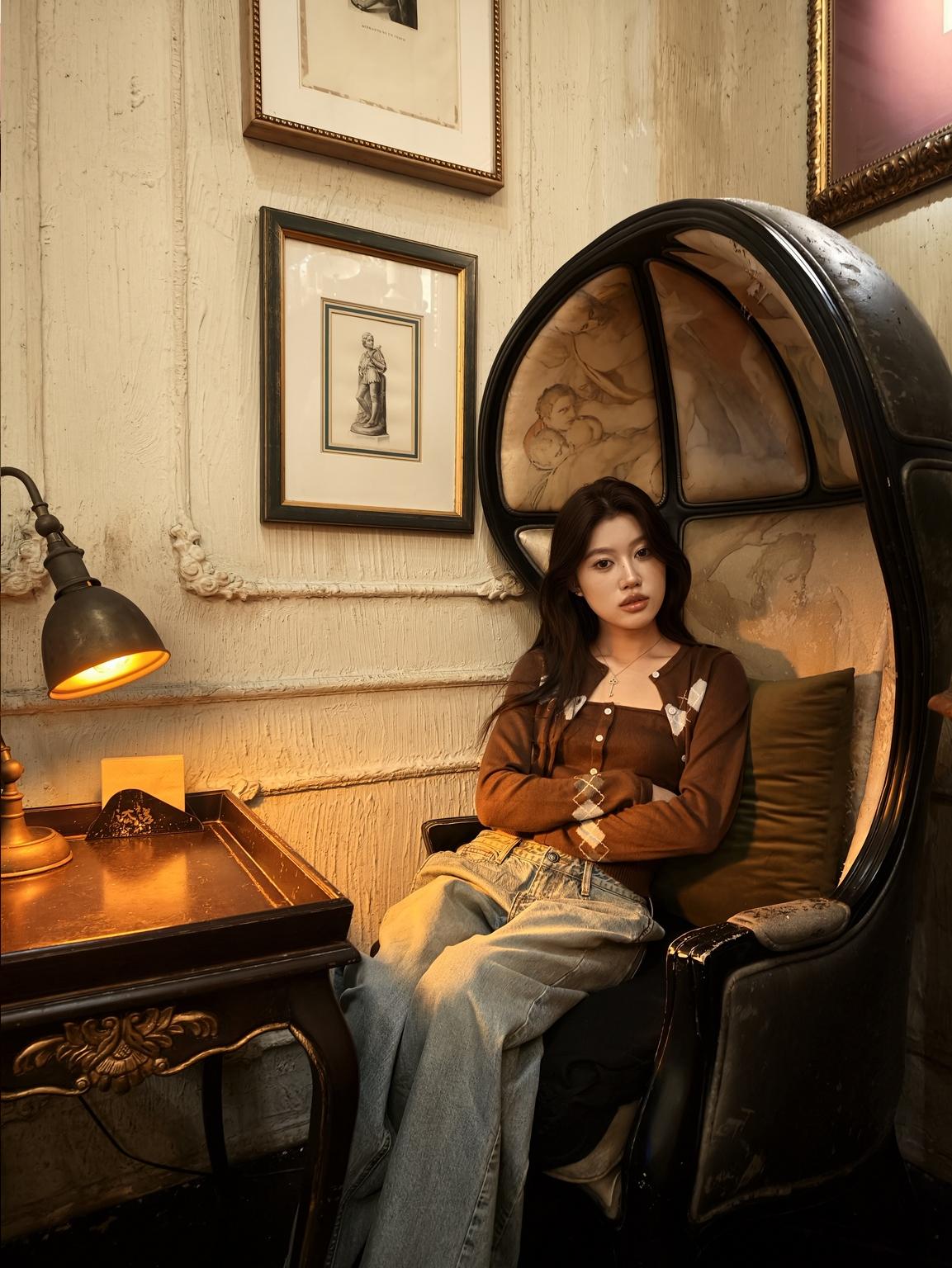} \\
\textbf{\tiny C} &
\includegraphics[width=\linewidth]{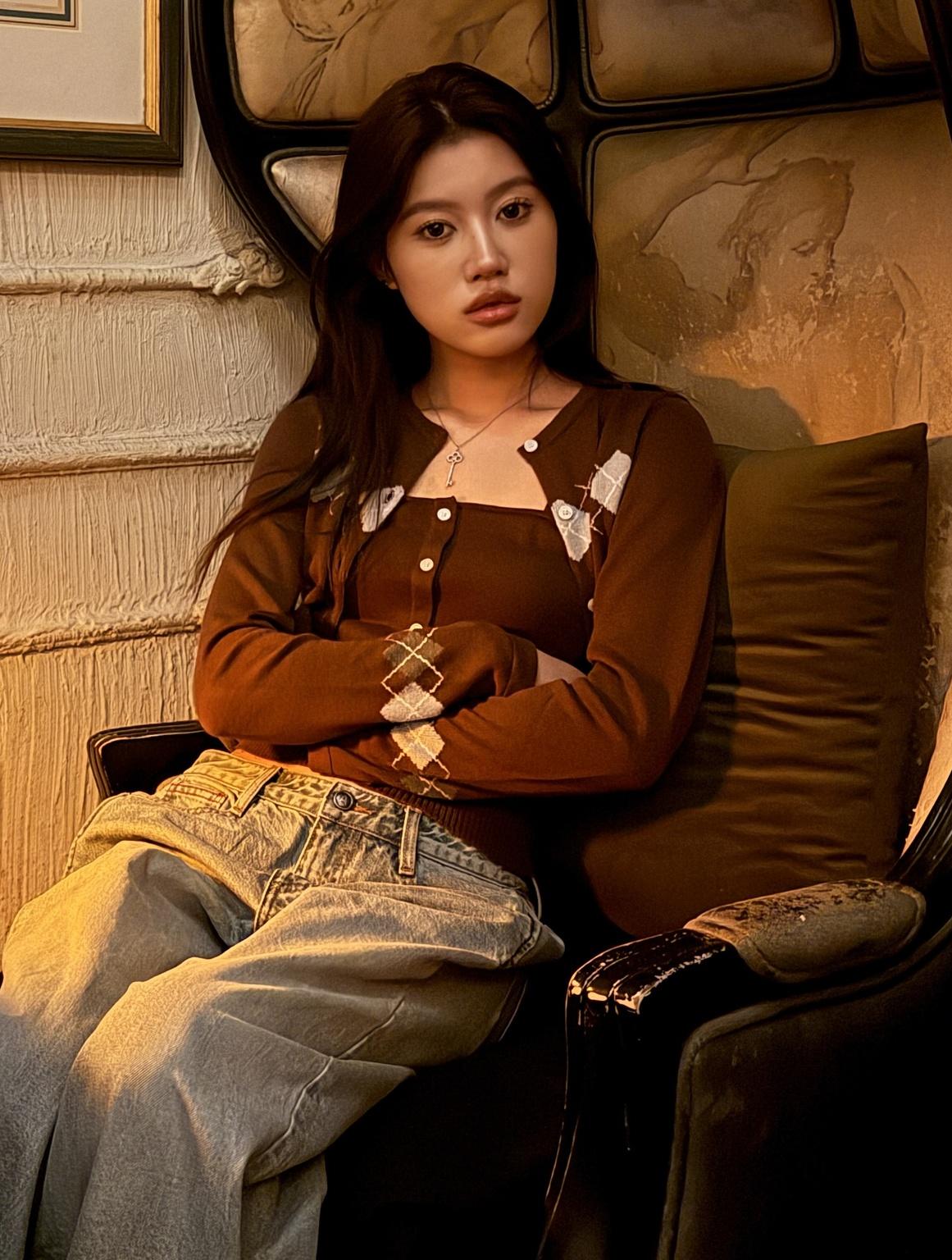} &
\includegraphics[width=\linewidth]{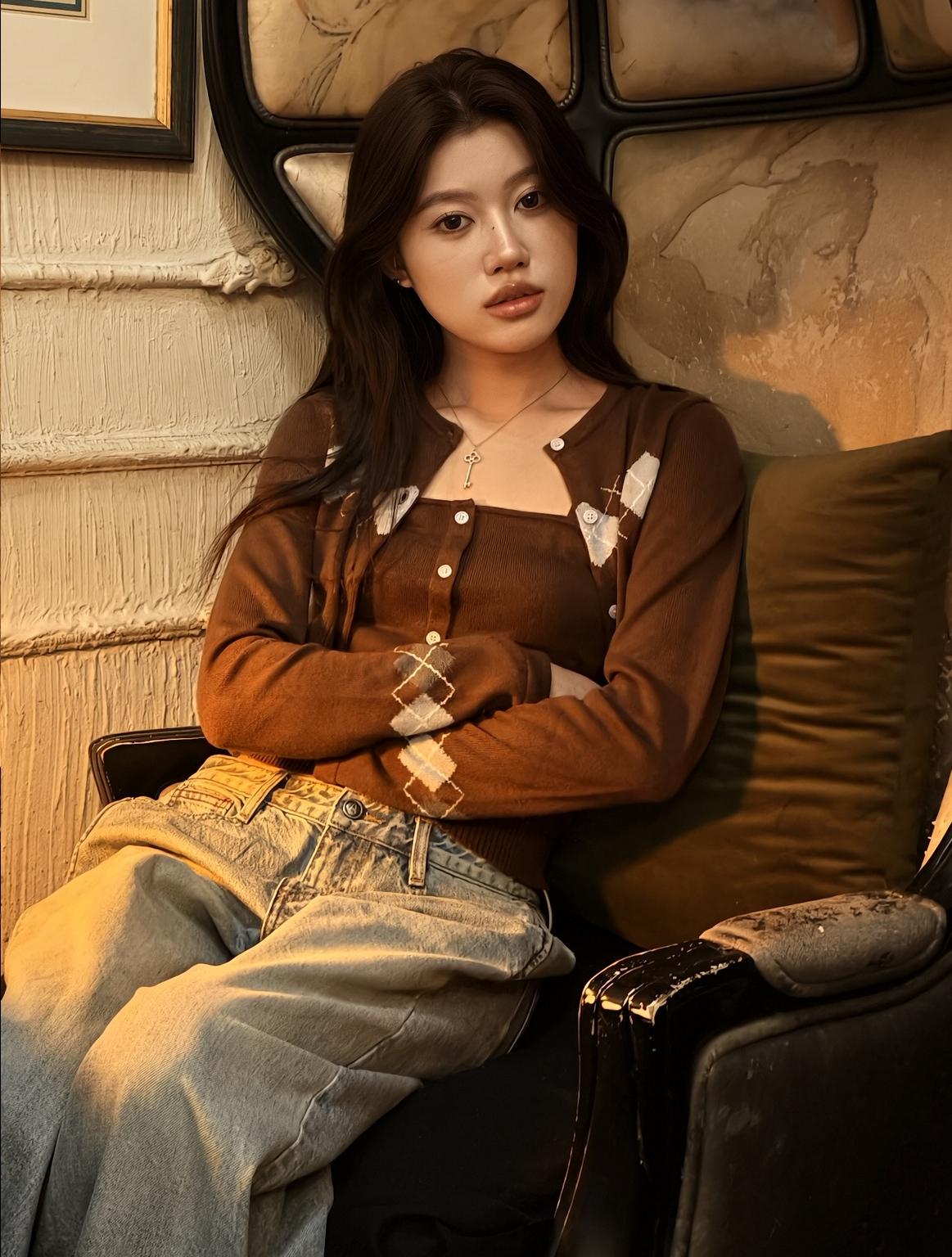}
\end{tabular}
&
\vcell{B / B / A / B / A}
&
\vcell{\textbf{B}}
&
Face Sharpness: Image B (Right) exhibits superior sharpness in facial details, particularly in the eyes and eyebrows, compared to Image A (Left), which appears softer. Face Texture: Image B (Right) demonstrates better texture rendering in the face, with more natural and fine details, whereas Image A (Left) shows less defined texture. Face Noise: Image B (Right) has lower noise levels in the facial region, resulting in a cleaner appearance, while Image A (Left) exhibits more noticeable noise.Global Sharpness: Image B (Right) maintains higher overall sharpness across the entire image, contributing to better clarity and detail retention.
\\
\midrule

\vcell{Failure}
&
\begin{tabular}{@{} c >{\centering\arraybackslash}m{0.13\textwidth} >{\centering\arraybackslash}m{0.13\textwidth} @{}}
& \textbf{\tiny A} & \textbf{\tiny B} \\
\textbf{\tiny G} &
\includegraphics[width=\linewidth]{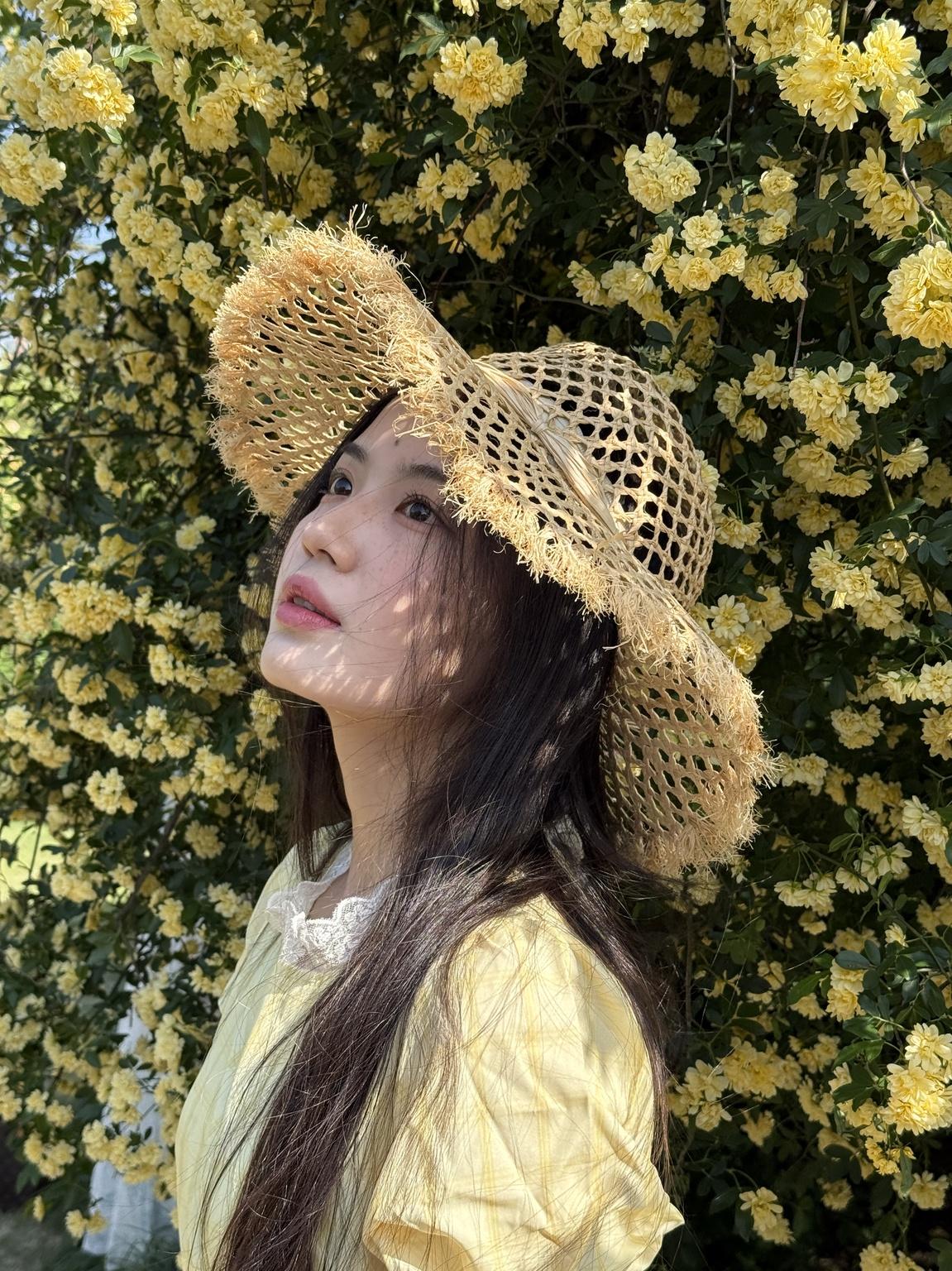} &
\includegraphics[width=\linewidth]{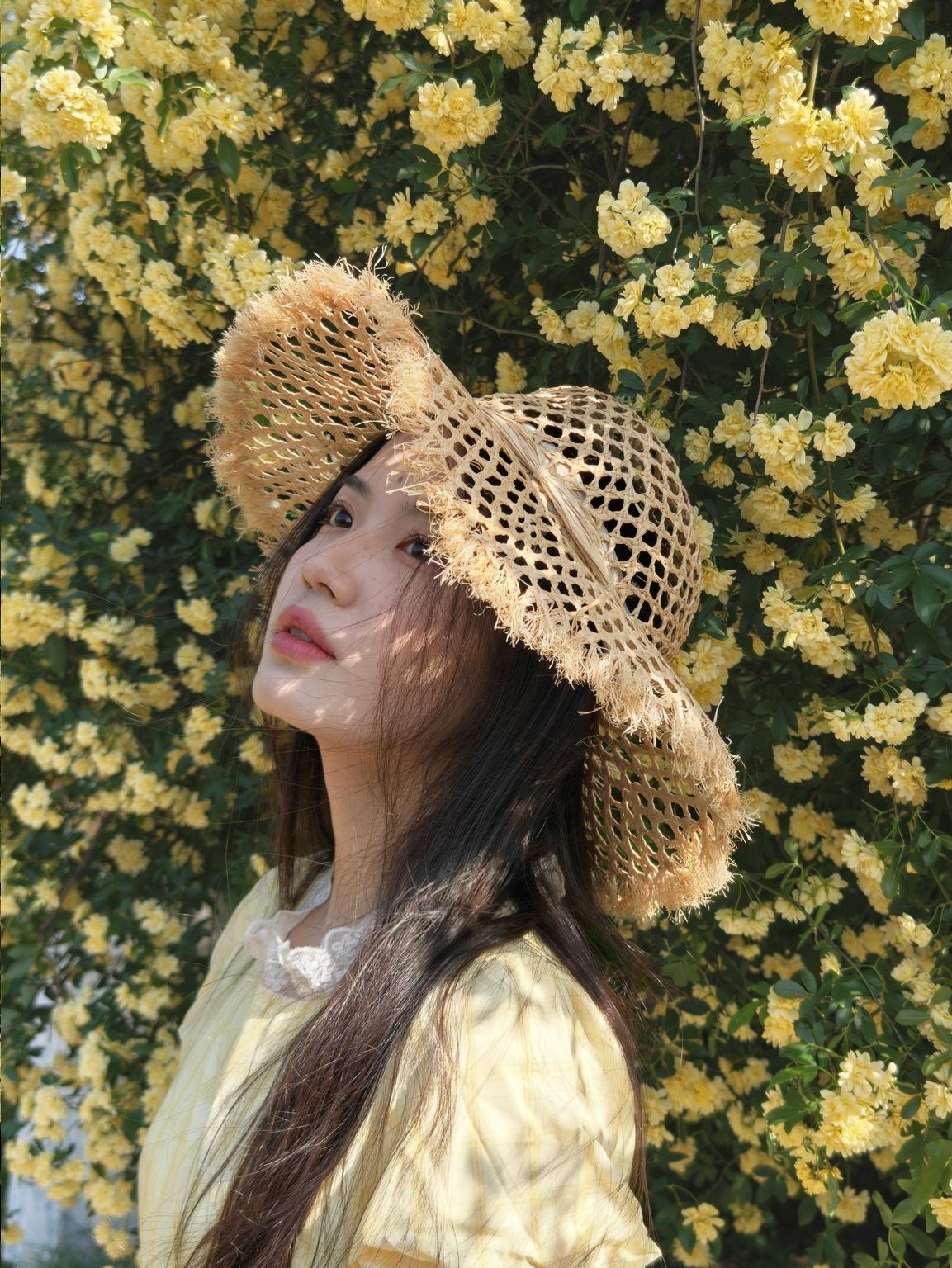} \\
\textbf{\tiny C} &
\includegraphics[width=\linewidth]{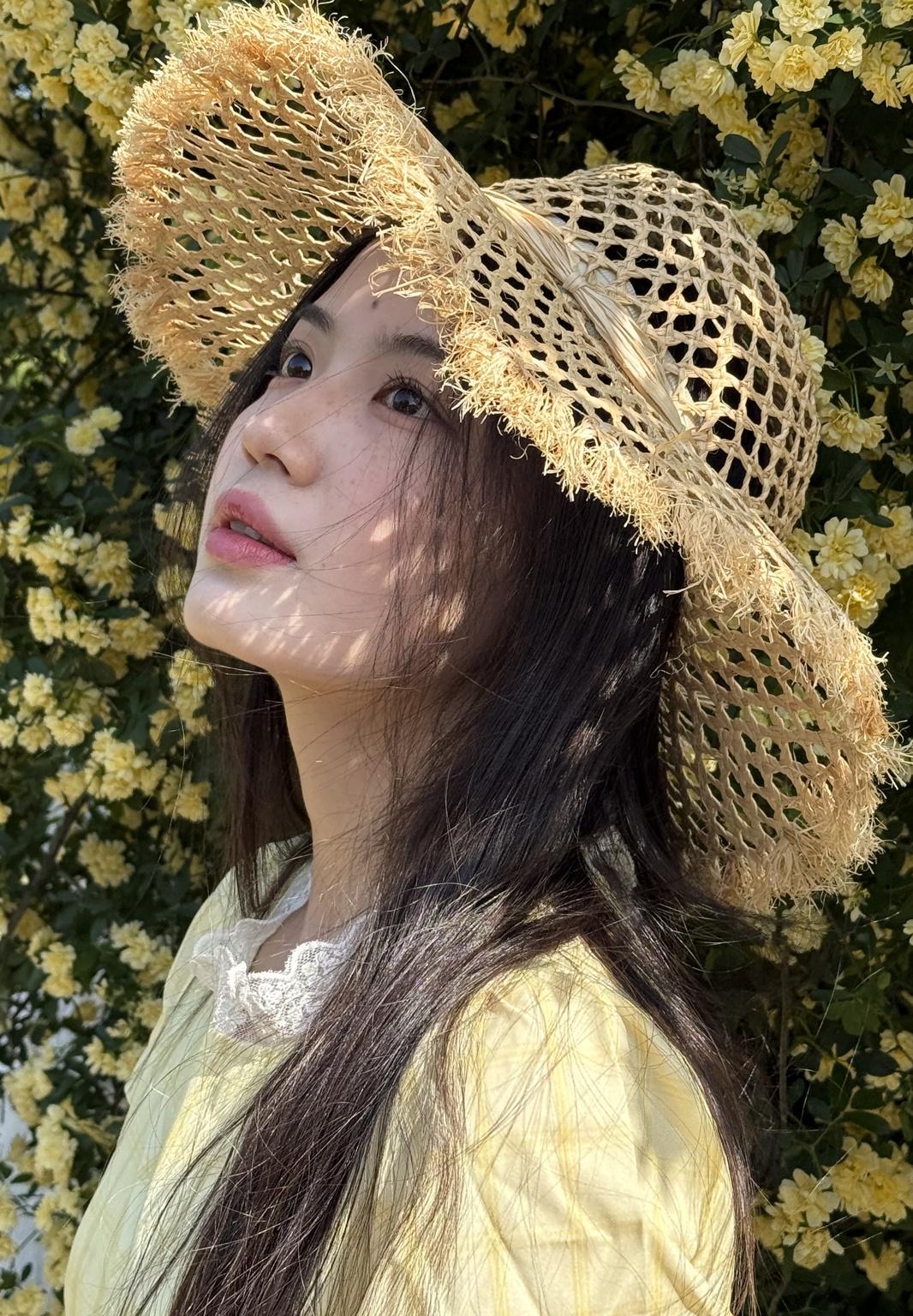} &
\includegraphics[width=\linewidth]{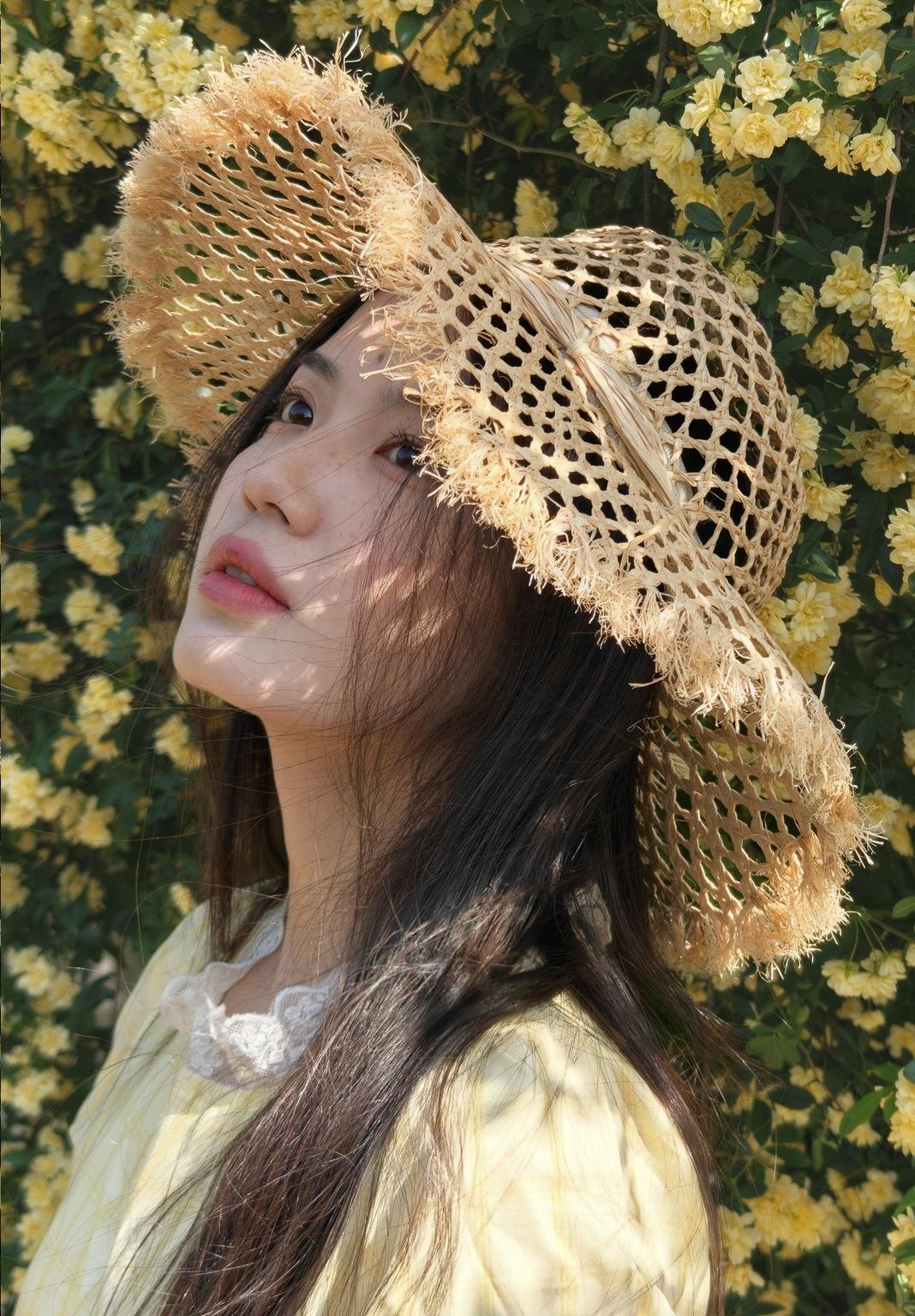}
\end{tabular}
&
\vcell{A / B / A / B / A}
&
\vcell{\textbf{A}}
&
Face Sharpness: Image B (Right) exhibits noticeable smearing in facial details, particularly in the eyes and eyebrows, while Image A (Left) maintains clearer and more defined features. Face Texture: Image B (Right) shows significant loss of fine texture in the facial region, resulting in a less detailed appearance, whereas Image A (Left) preserves more natural skin texture. Hair Texture: Image B (Right) demonstrates severe smearing in the hair strands, leading to a loss of individual hair detail, while Image A (Left) retains better hair texture and definition. Global Sharpness: Image B (Right) suffers from overall blurriness and smearing, especially in high-frequency areas like hair and facial details, while Image A (Left) presents a sharper and more detailed image.
\\
\bottomrule[1.0pt]
\end{tabularx}
}
\end{table*}

\begin{table*}[t]
\centering
\scriptsize
\setlength{\tabcolsep}{2pt}
\renewcommand{\arraystretch}{1.0}
\caption{Qualitative examples on the scene category of the proposed iDiff pipeline.
For each case, we show the four input images arranged in a 2$\times$2 layout, the predictions of five Answer Models, the final voting result, and the reasoning generated by the Thinking Model.}
\label{tab:qualitative_cases_land}
\resizebox{\textwidth}{!}{
\begin{tabularx}{\textwidth}
{
>{\centering\arraybackslash}m{0.05\textwidth}
>{\centering\arraybackslash}m{0.45\textwidth}
>{\centering\arraybackslash}m{0.10\textwidth}
>{\centering\arraybackslash}m{0.05\textwidth}
>{\raggedright\arraybackslash}X
}
\toprule[1.0pt]
\textbf{Case} & \textbf{Inputs} & \textbf{Answer Models} & \textbf{Vote} & \theadcenter{Thinking Model}\\
\midrule

\vcell{Success}
&
\begin{tabular}{@{} c >{\centering\arraybackslash}m{0.13\textwidth} >{\centering\arraybackslash}m{0.13\textwidth} @{}}
& \textbf{\tiny A} & \textbf{\tiny B} \\
\textbf{\tiny G} &
\includegraphics[width=\linewidth]{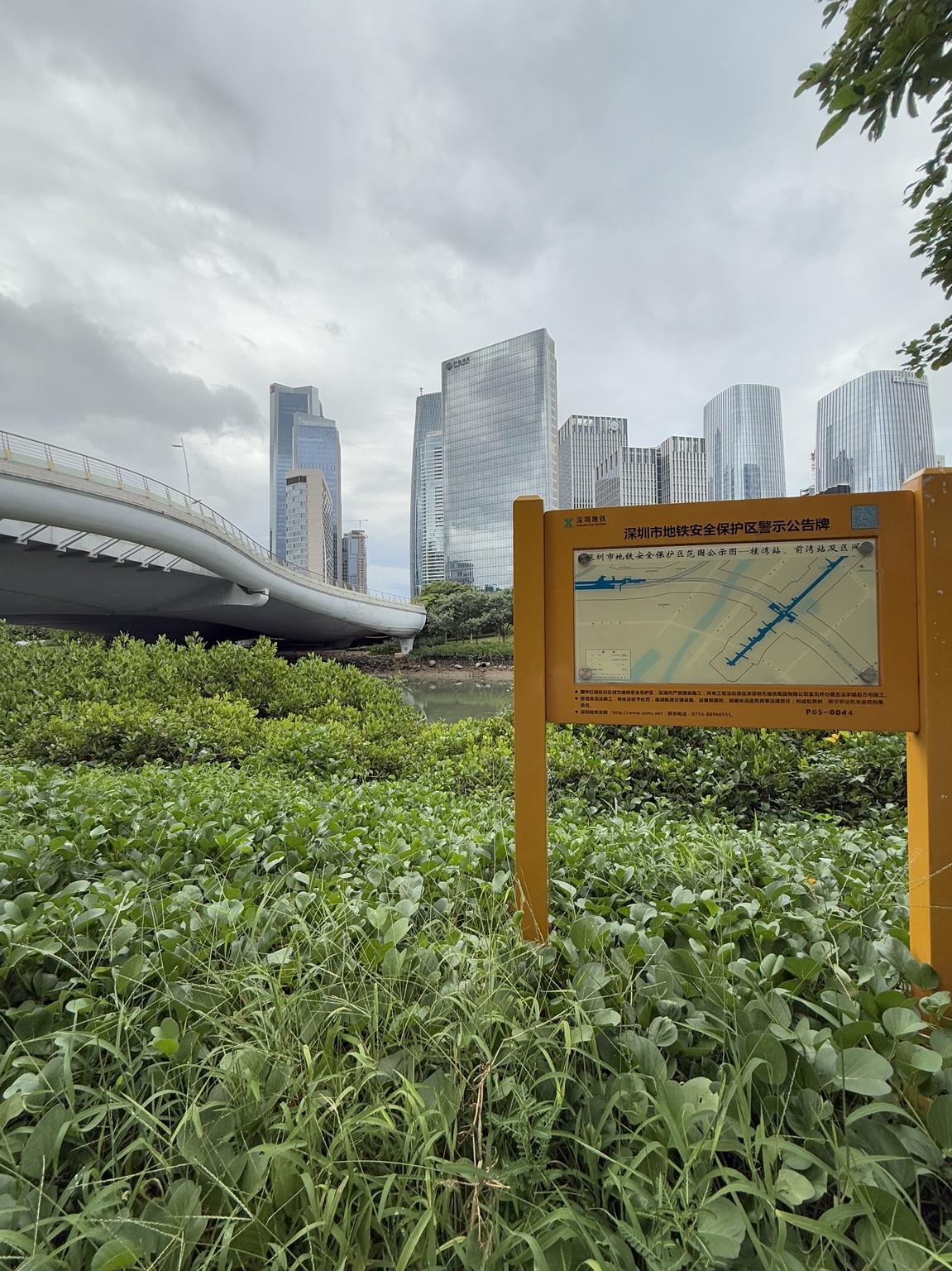} &
\includegraphics[width=\linewidth]{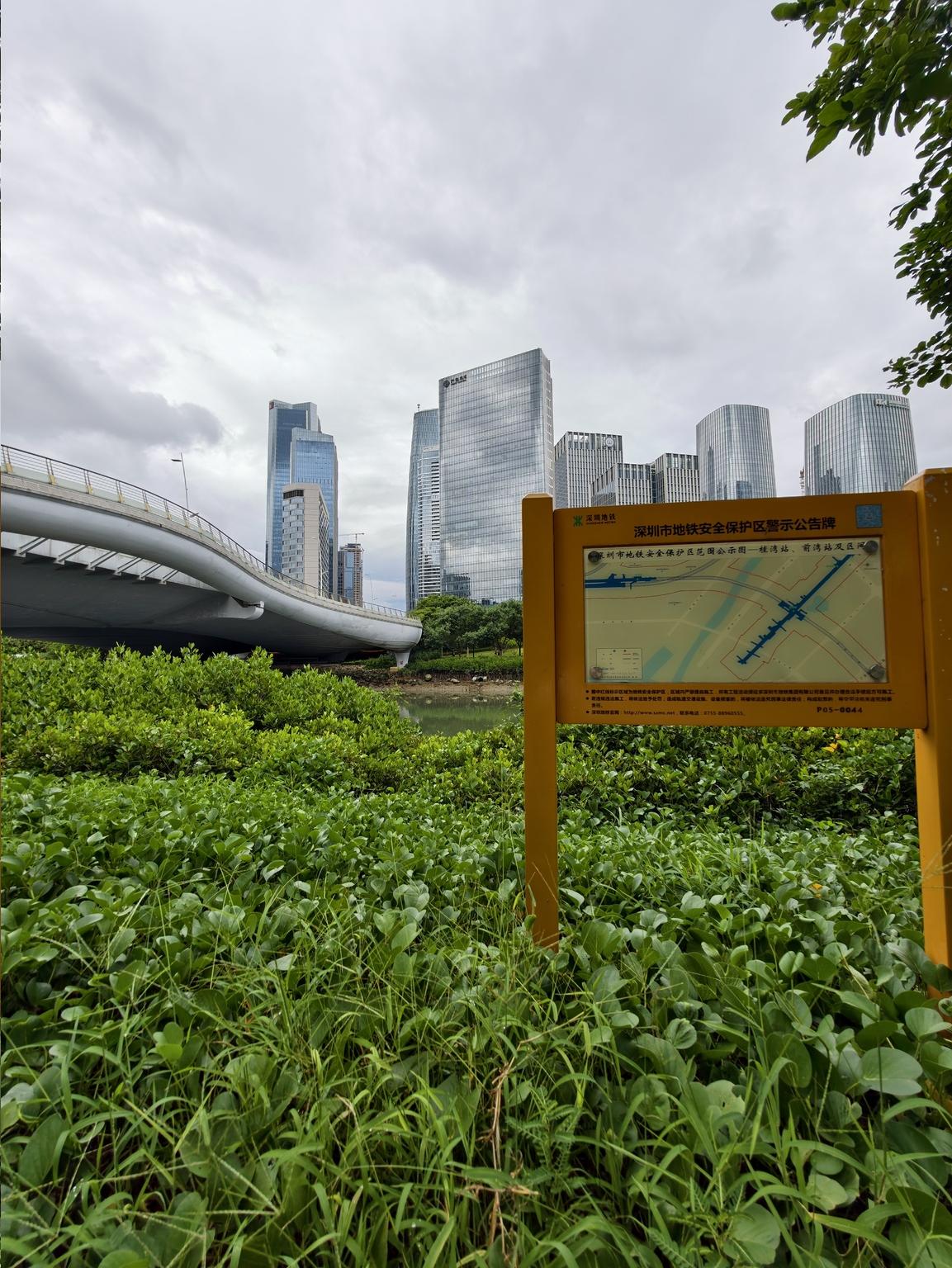} \\
\textbf{\tiny C} &
\includegraphics[width=\linewidth]{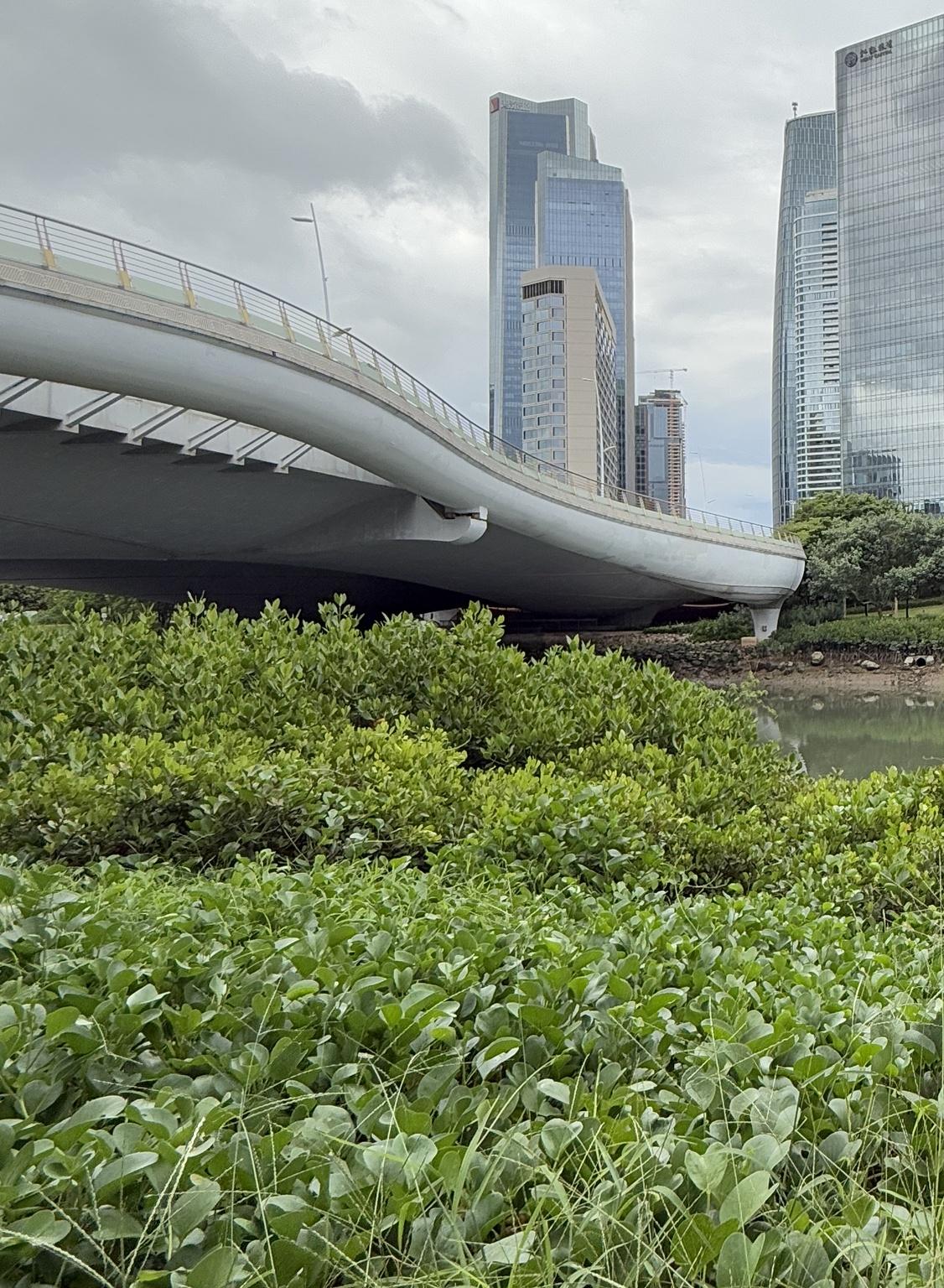} &
\includegraphics[width=\linewidth]{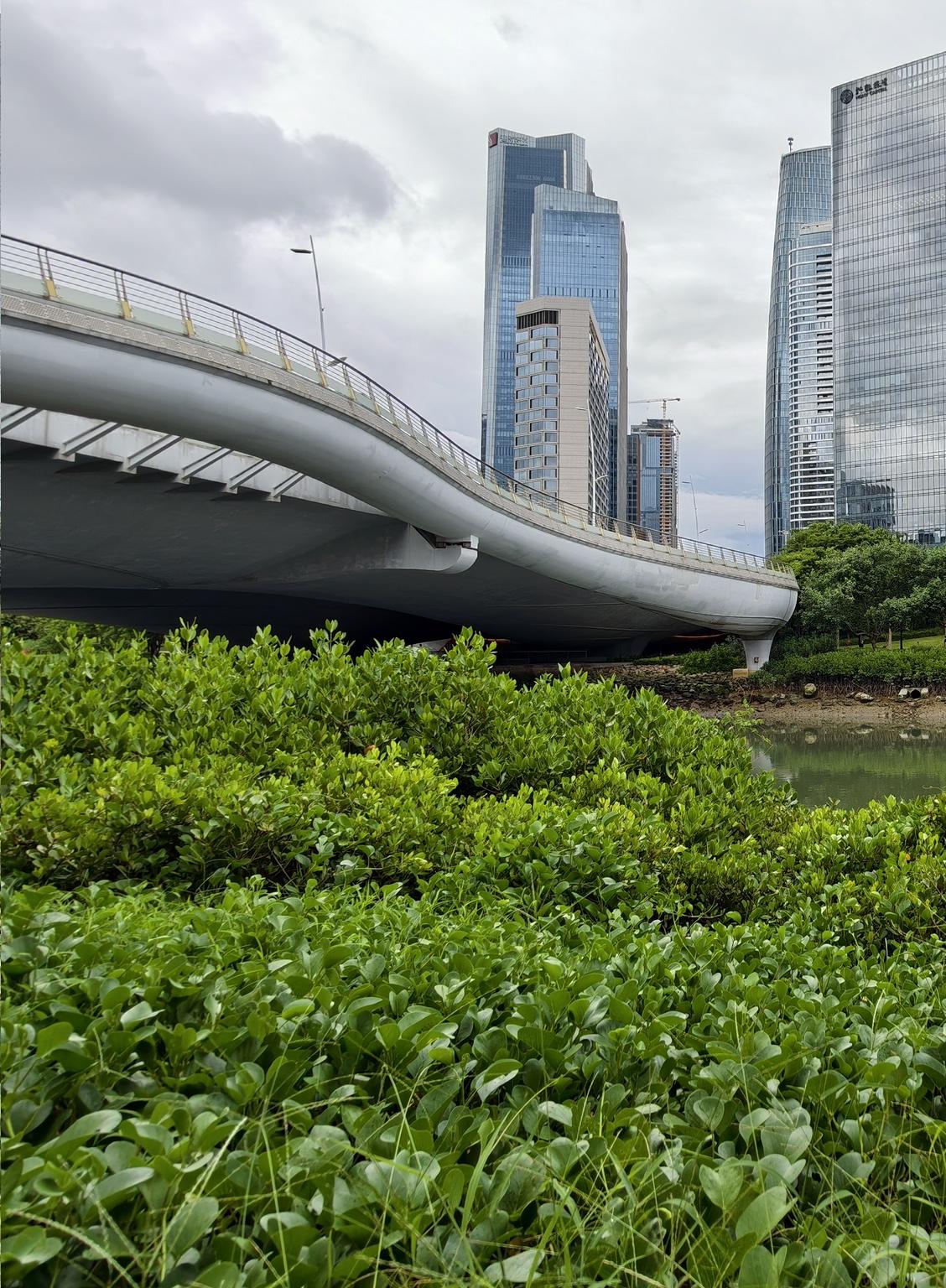}
\end{tabular}
&
\vcell{B / B / B / B / B}
&
\vcell{\textbf{B}}
&
Global Sharpness: Image B (Right) exhibits superior overall sharpness and clarity, particularly in fine details such as the texture of the grass and the edges of the signboard, compared to Image A (Left) which appears softer. Global Noise: Image B (Right) demonstrates better noise control, with less visible noise in mid-to-dark regions, resulting in a cleaner and more natural appearance. Global Texture: Image B (Right) preserves finer texture details in both the foreground vegetation and the distant buildings, contributing to a more realistic and detailed image.
\\
\midrule

\vcell{Challenging}
&
\begin{tabular}{@{} c >{\centering\arraybackslash}m{0.13\textwidth} >{\centering\arraybackslash}m{0.13\textwidth} @{}}
& \textbf{\tiny A} & \textbf{\tiny B} \\
\textbf{\tiny G} &
\includegraphics[width=\linewidth]{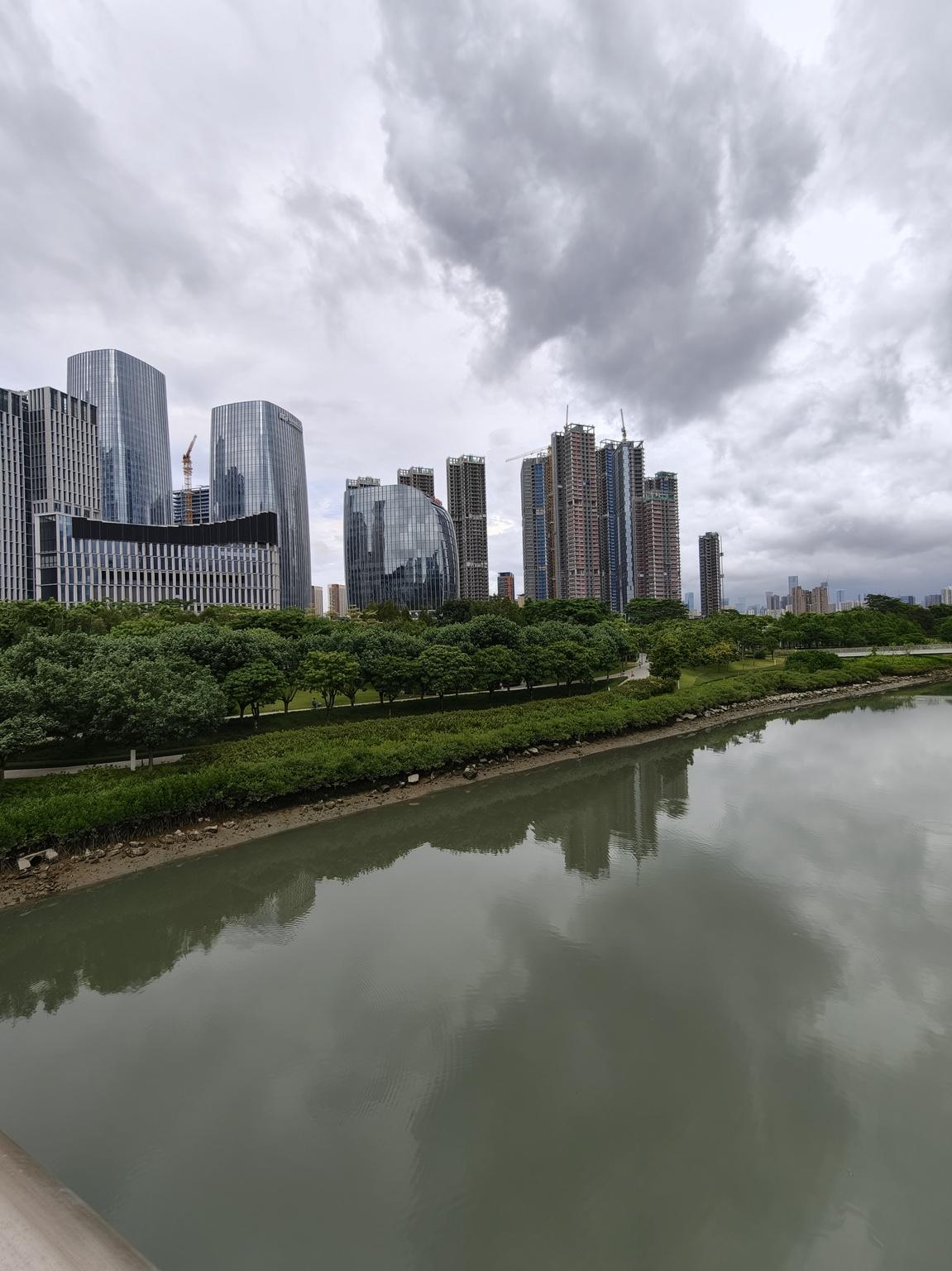} &
\includegraphics[width=\linewidth]{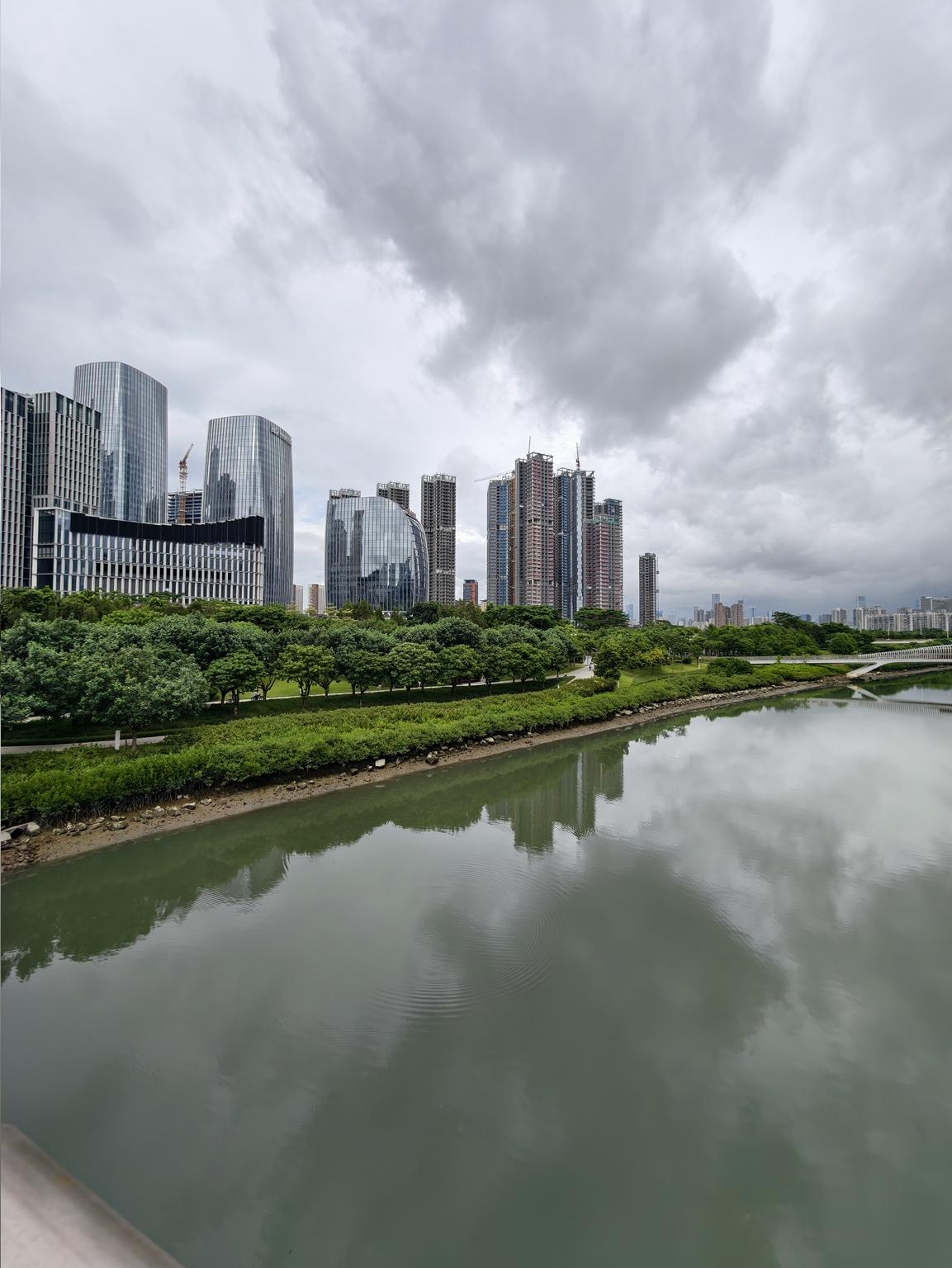} \\
\textbf{\tiny C} &
\includegraphics[width=\linewidth]{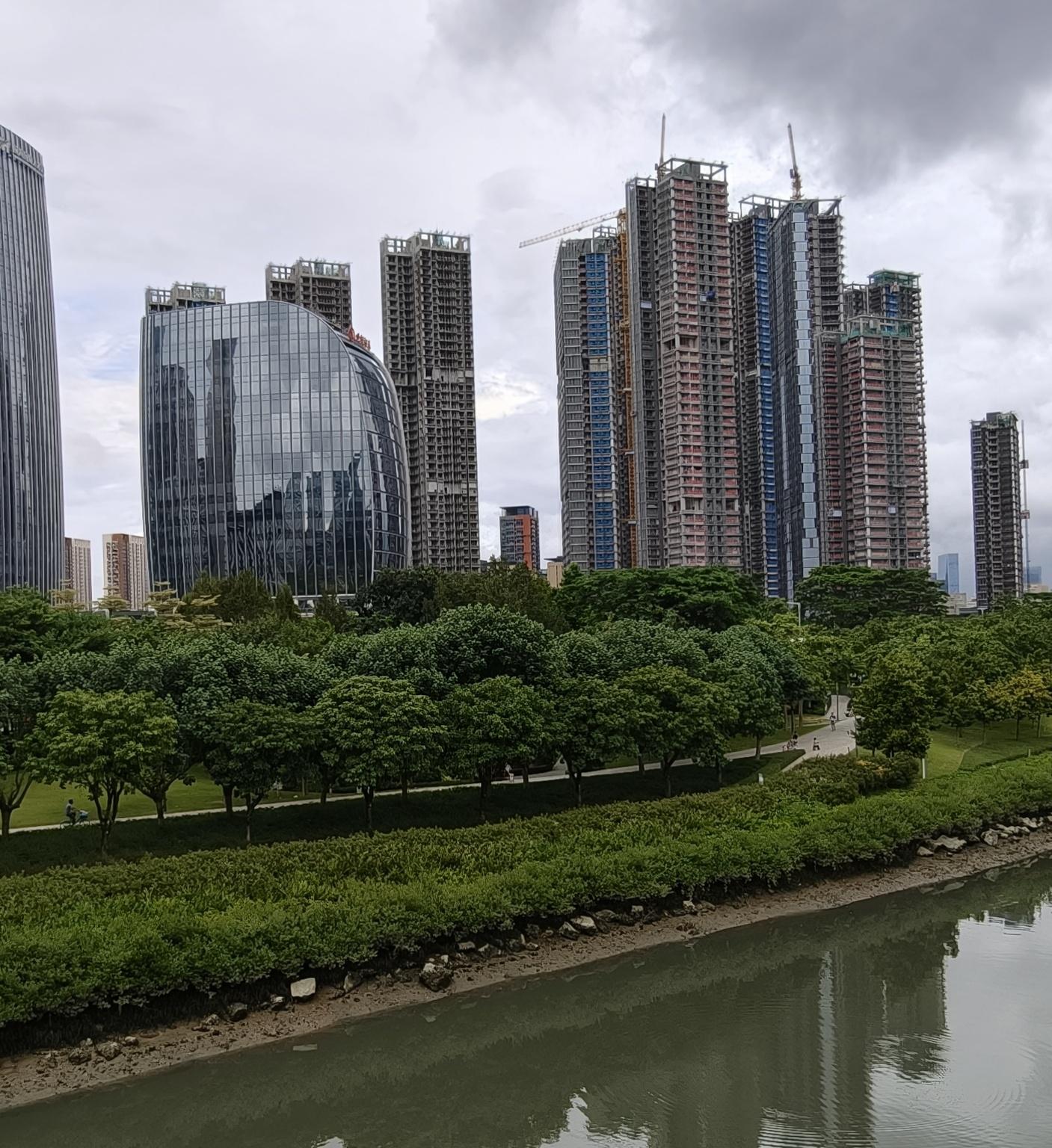} &
\includegraphics[width=\linewidth]{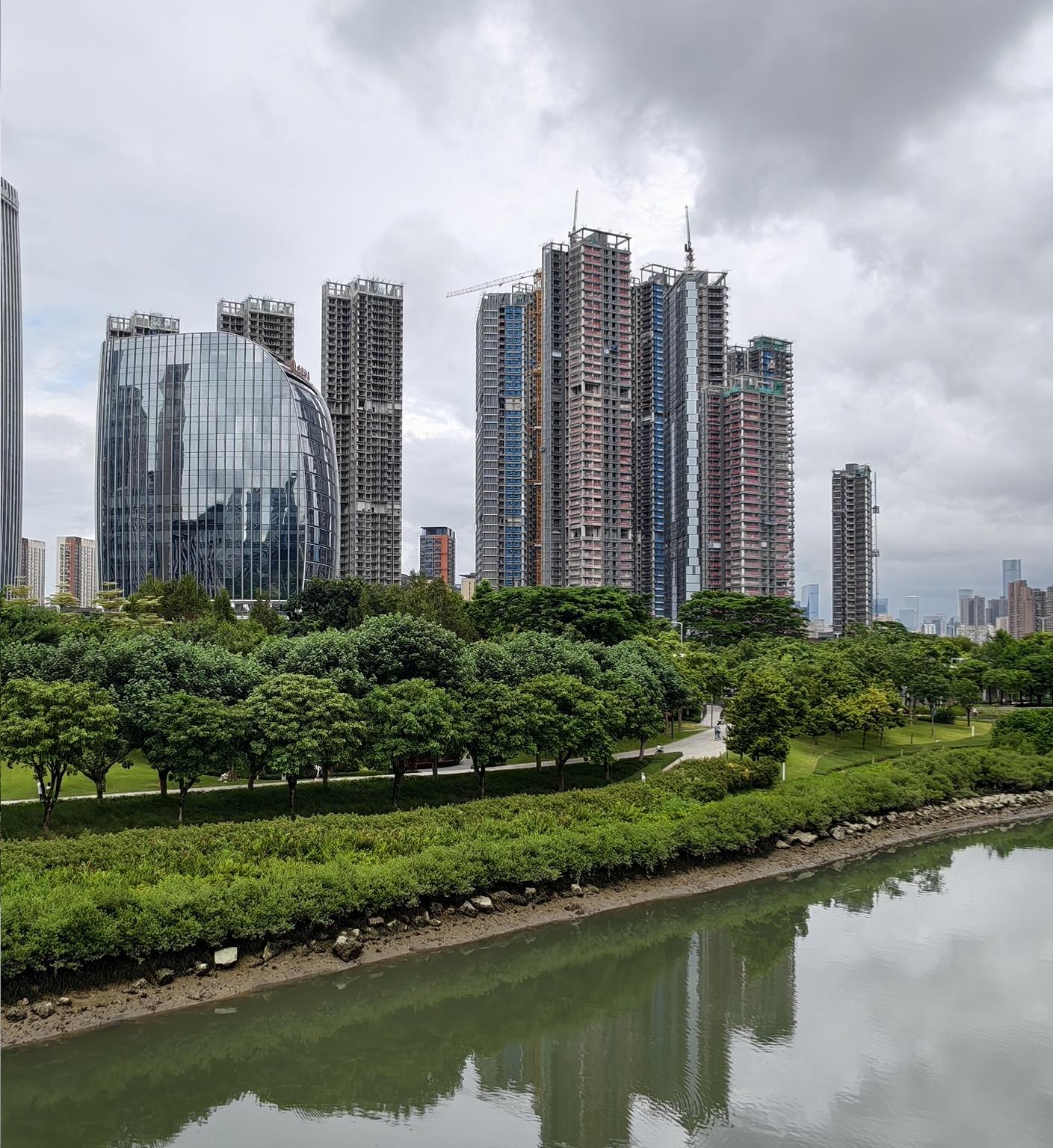}
\end{tabular}
&
\vcell{A / B / A / B / B}
&
\vcell{\textbf{B}}
&
Global Sharpness: Image B (Right) exhibits superior overall sharpness and clarity, particularly in fine details such as the texture of the water surface and distant buildings, whereas Image A (Left) appears softer and less defined. Global Noise: Image B (Right) demonstrates better noise control, with less visible grain or smearing in mid-to-dark regions, contributing to a cleaner and more natural appearance. Global Texture: Image B (Right) preserves finer textures across the entire scene, including the water ripples and building facades, while Image A (Left) shows a loss of detail in these areas.
\\
\midrule

\vcell{Failure}
&
\begin{tabular}{@{} c >{\centering\arraybackslash}m{0.13\textwidth} >{\centering\arraybackslash}m{0.13\textwidth} @{}}
& \textbf{\tiny A} & \textbf{\tiny B} \\
\textbf{\tiny G} &
\includegraphics[width=\linewidth]{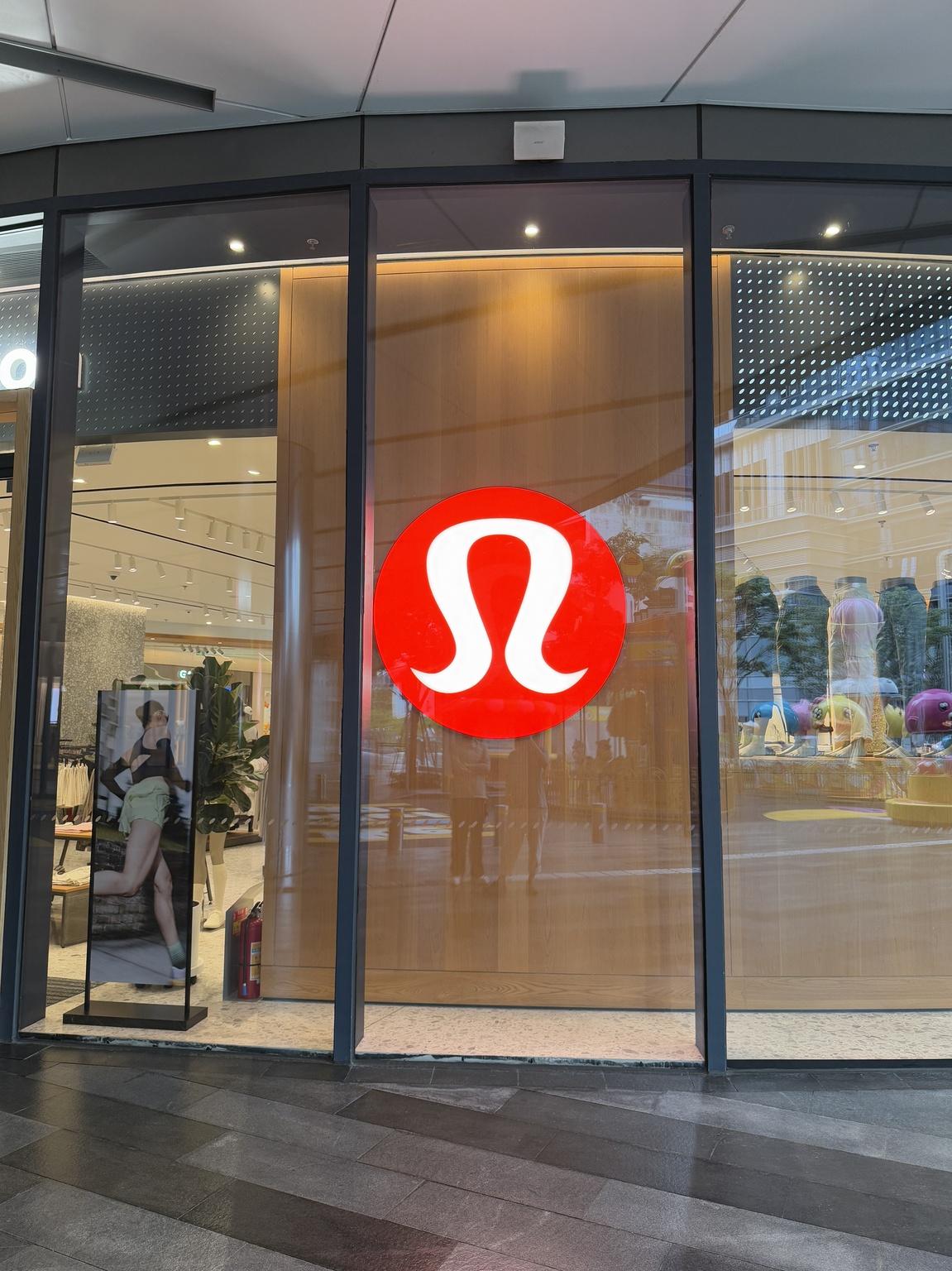} &
\includegraphics[width=\linewidth]{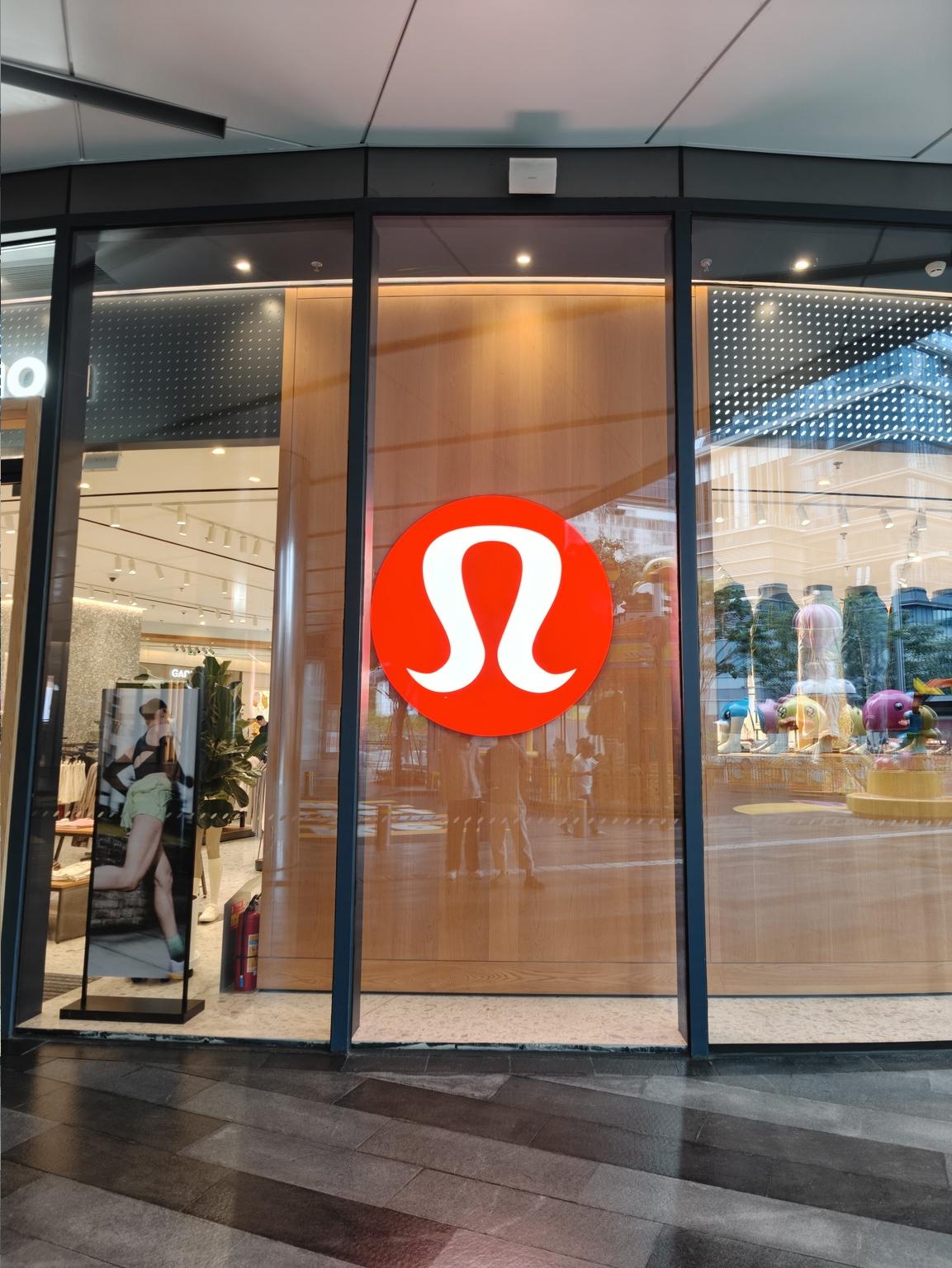} \\
\textbf{\tiny C} &
\includegraphics[width=\linewidth]{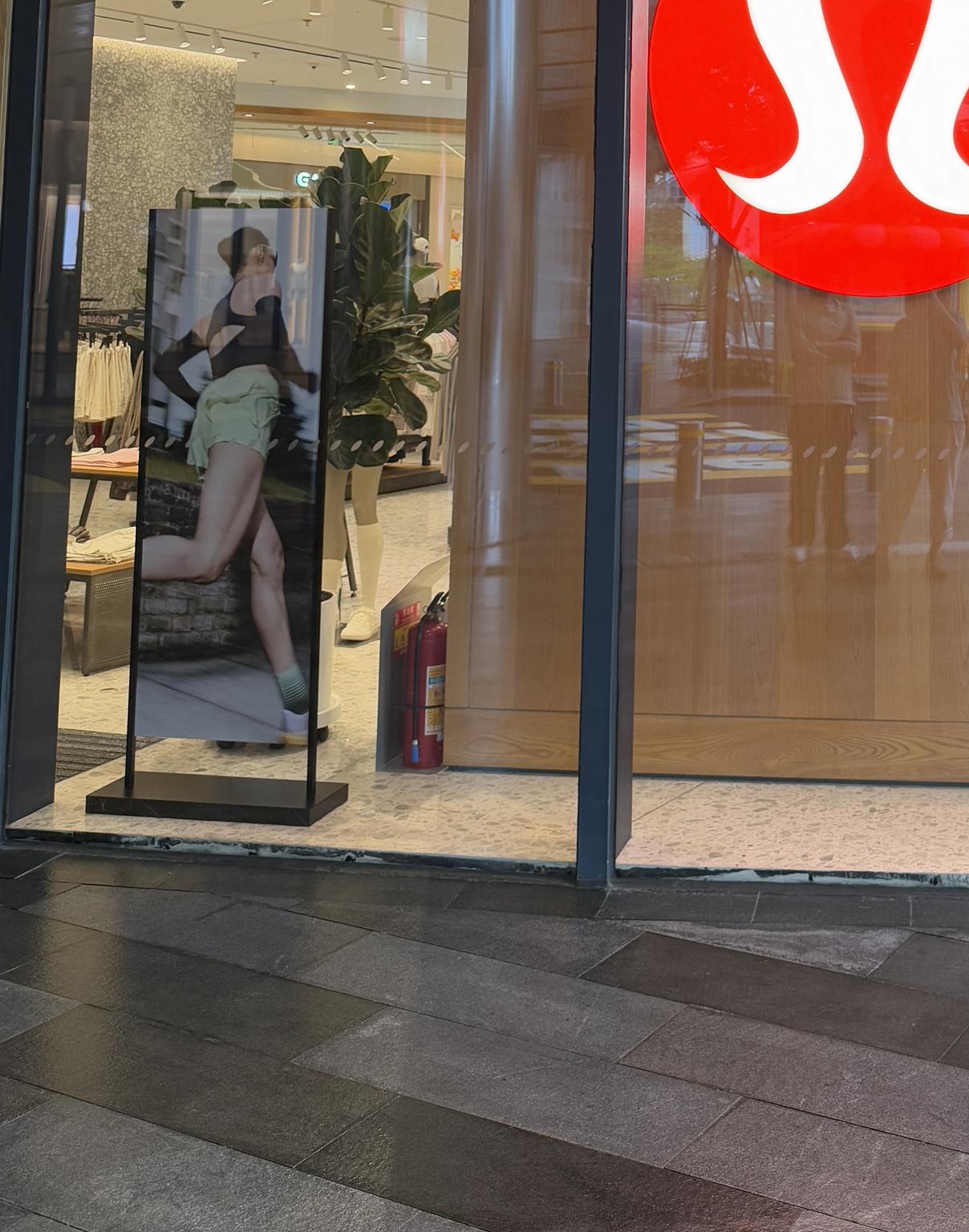} &
\includegraphics[width=\linewidth]{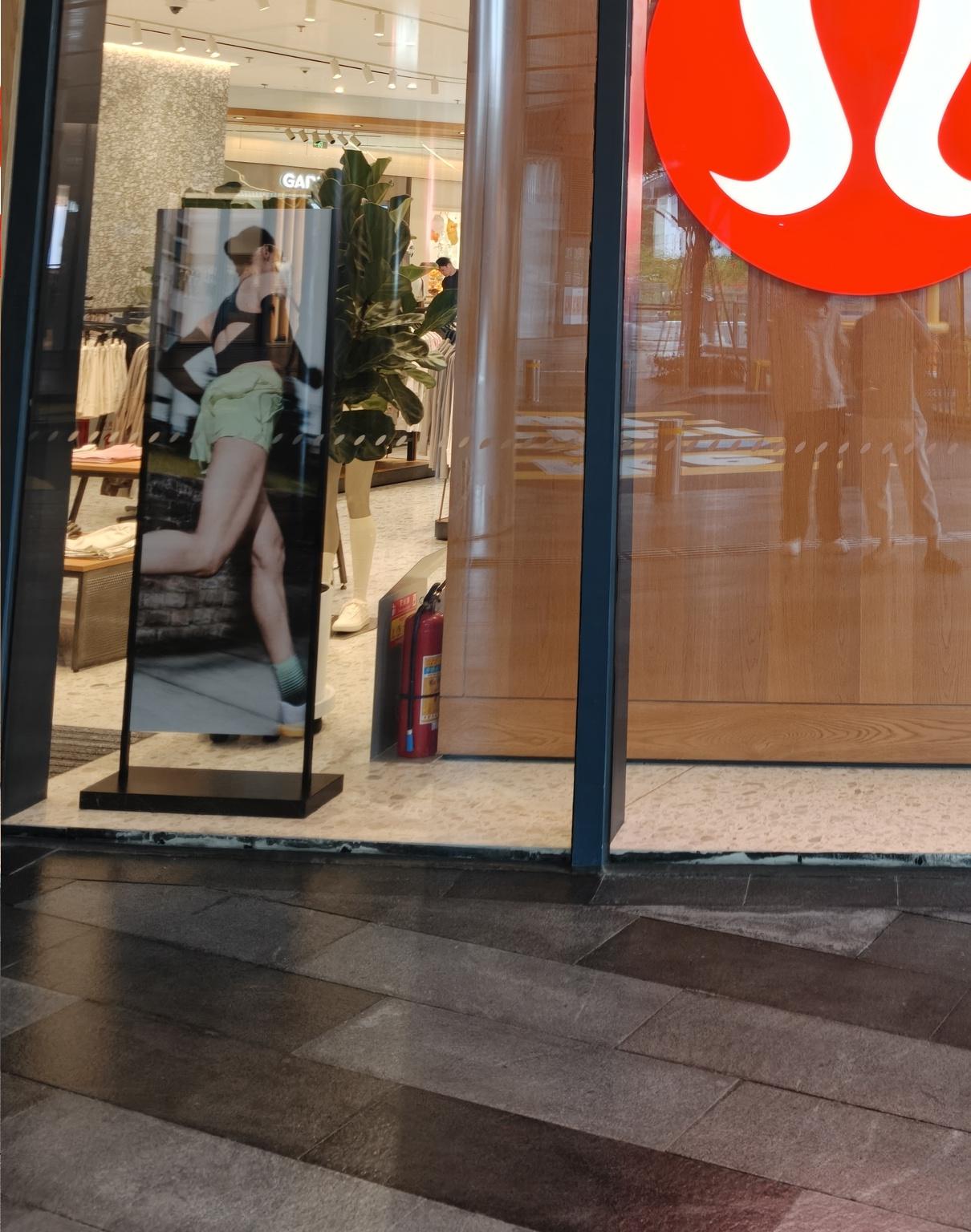}
\end{tabular}
&
\vcell{A / B / A / B / B}
&
\vcell{\textbf{B}}
&
Global Sharpness: Image B (Right) exhibits superior overall sharpness and clarity, particularly in fine details such as the texture of the glass and the edges of the logo, while Image A (Left) appears softer and less defined. Global Noise: Image B (Right) shows more pronounced noise in darker areas, which is a common artifact in high-contrast scenes, whereas Image A (Left) has less visible noise but at the cost of detail. Global Artifacts: Image B (Right) displays noticeable black borders along the edges of the glass panels, which may indicate over-sharpening or edge enhancement artifacts, while Image A (Left) lacks these specific artifacts but appears less detailed overall.
\\
\bottomrule[1.0pt]
\end{tabularx}
}
\end{table*}

\end{document}